\documentclass[10pt,twocolumn,letterpaper]{article}
\usepackage[pagenumbers]{cvpr} % To force page numbers, e.g. for an arXiv version
\definecolor{cvprblue}{rgb}{0.21,0.49,0.74}
\usepackage[pagebackref,breaklinks,colorlinks,allcolors=cvprblue]{hyperref}

\usepackage[ruled,vlined,linesnumbered]{algorithm2e}
\usepackage{amsmath}
\usepackage{amssymb}
\usepackage{tikz}
\usetikzlibrary{shapes,arrows,positioning}
\usepackage{dirtree}
\usepackage{ifthen} % For conditional logic
\usepackage{booktabs} % For table
\usepackage{multirow} % For table
\usepackage{float}
\usepackage{adjustbox}
\usepackage[table]{xcolor}
\usepackage{colortbl}
\definecolor{midgray}{gray}{0.8}
\definecolor{lightgray}{gray}{0.9}
\definecolor{r}{rgb}{0.8,0.0,0.0}
\definecolor{g}{rgb}{0.0,0.5,0.0}
\definecolor{sectiongray}{gray}{0.90}

\usepackage[accsupp]{axessibility}  % Improves PDF readability for those with disabilities.

\title{Toward Generalizable Whole Brain Representations\\with High-Resolution Light-Sheet Data}

\author{
Minyoung E. Kim$^{*,\dagger}$
\hspace{0.8em}
Dae Hee Yun\footnotemark[1]
\hspace{0.8em}
Aditi V. Patel
\hspace{0.8em}
Madeline Hon \\
Webster Guan
\hspace{0.8em}
Taegeon Lee
\hspace{0.8em}
Brian Nguyen \\
LifeCanvas Technologies\\
{\tt\small \{mykim, daeheeyun, aditi, mhon, websterg, t76lee, brian\}@lifecanvastech.com}
}
\makeatletter
\g@addto@macro\@maketitle{\@thanks} % <-- title footnotes(=thanks) actually get printed
\makeatother

\begin{document}
%\maketitle

\twocolumn[{%
\renewcommand\twocolumn[1][]{#1}%
\maketitle
\begin{center}
    \centering
    \captionsetup{type=figure}
    \includegraphics[width=\textwidth,height=5cm]{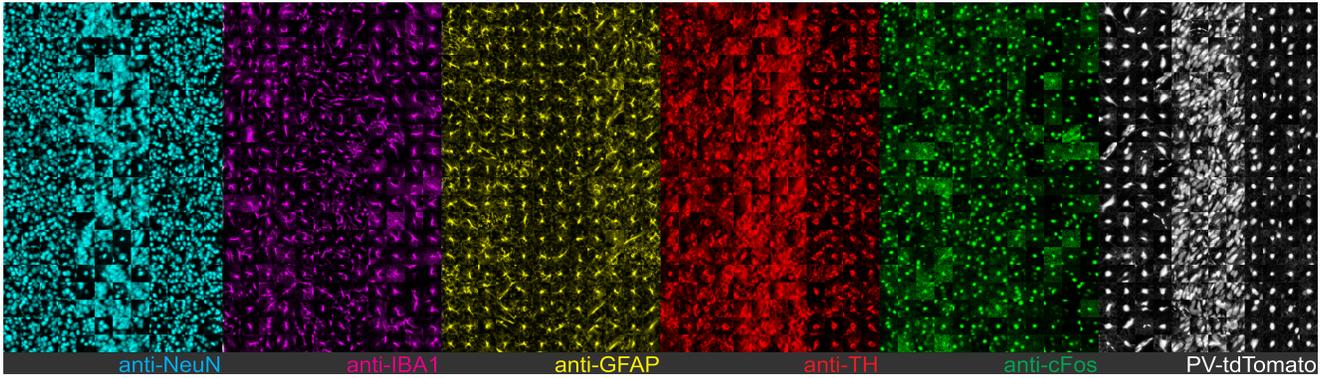}
    \captionof{figure}{CANVAS: A high-resolution light-sheet benchmark dataset, representing 6 different cell types at subcellular resolution.}
    \label{fig:canvas}
    \vspace{10pt}
\end{center}%
}]

\BlankLine

\begingroup
\renewcommand\thefootnote{\fnsymbol{footnote}}
\footnotetext[1]{Equal contribution}
\footnotetext[2]{Corresponding author}
\endgroup

\begin{abstract}
Unprecedented visual details of biological structures are being revealed by subcellular-resolution whole-brain 3D microscopy data, enabled by recent advances in intact tissue processing and light-sheet fluorescence microscopy (LSFM). These volumetric data offer rich morphological and spatial cellular information, however, the lack of scalable data processing and analysis methods tailored to these petabyte-scale data poses a substantial challenge for accurate interpretation. Further, existing models for visual tasks such as object detection and classification struggle to generalize to this type of data. To accelerate the development of suitable methods and foundational models, we present CANVAS, a comprehensive set of high-resolution whole mouse brain LSFM benchmark data, encompassing six neuronal and immune cell-type markers, along with cell annotations and a leaderboard. We also demonstrate challenges in generalization of baseline models built on existing architectures, especially due to the heterogeneity in cellular morphology across phenotypes and anatomical locations in the brain. To the best of our knowledge, CANVAS is the first and largest LSFM benchmark that captures intact mouse brain tissue at subcellular level, and includes extensive annotations of cells throughout the brain.
\end{abstract}
\vspace{-7pt} 
\section{Introduction}
\label{sec:intro}
The brain is a complex organ comprising thousands of distinct cell types with diverse molecular, morphological, and functional properties. These cellular phenotypes often exhibit region-specific profiles and heterogeneous distributions across the brain \cite{Bryant2023-wn,Das2023-yk,Miller_2018}, reflecting local microenvironmental factors. Thin-section imaging can only provide a limited view of cellular diversity \cite{Ueda2020-tk} and cannot be reliably extrapolated to capture holistic cell profiles across the whole brain. Moreover, comprehensive characterization of cell structures requires imaging approaches that achieve subcellular resolution. Recent advances in light-sheet fluorescence microscopy (LSFM), together with tissue clearing and uniform immunohistochemistry (IHC) labeling techniques, have enabled high-resolution imaging of intact tissues, including the whole mouse brain. Such advances open new opportunities for in-depth phenotyping of individual cells throughout the brain, integrating their molecular, morphological, and microenvironmental features.

However, large-scale, subcellular-resolution 3D data of intact tissues, such as mouse brains, often reach the petabyte scale in real-world applications, and the lack of robust data processing and analysis techniques tailored to these datasets hinders comprehensive interpretation. Scalable end-to-end ETL (Extract, Transform, Load) pipelines and generalizable machine learning methods are essential for extracting meaningful biological insights. While machine learning has made remarkable progress in computer vision tasks including object detection and classification, existing methods in biomedical imaging and related domains still struggle to generalize to these types of data. As new data modalities continue to emerge, the need for foundational models that can capture generalizable features across modalities is becoming increasingly important. Although various strategies have been proposed to build such models without large annotated datasets, making both data and annotations publicly available remains critical for accelerating progress.

In this work, we present CANVAS, a new high-resolution light-sheet dataset and public leaderboard for whole-brain cell detection. The dataset consists of 3D LSFM volumes of intact mouse brains labeled with one of six cell-type markers (NeuN, cFos, PV, TH, Iba1, and GFAP), each targeting specific neuronal or immune populations (\Cref{fig:canvas}). The raw volumetric data (65–140 GB compressed) cover the entire mouse brain except for cFos, which is hemisphere-only, and contain 1,600–1,850 z-slices at approximately 7,000$\times$10,000 pixels per slice. Each marker dataset includes cell-centroid annotations for three regions of interest (\Cref{tab:annotation}) for model training and testing. These regions contain between 109 and 9,577 annotated cells, yielding 45,745 training and 47,301 test cell centroids in total. Using CANVAS, we show that baseline models built on existing architectures struggle to generalize across markers and regions, highlighting the need for publicly available datasets such as CANVAS and the domain's need for robust foundational models. To our knowledge, CANVAS represents the first and largest publicly available whole-brain LSFM dataset with extensive annotations, evaluation metrics, and a benchmark leaderboard, providing a critical resource for developing scalable and generalizable models for object detection in LSFM and other volumetric biological data. The dataset, annotations, and leaderboard are available at \href{https://canvas.lightsheetdata.com}{https://canvas.lightsheetdata.com}.

\section{Background and Motivation}
\label{sec:background}
Intact tissue 3D image data, integral to a holistic understanding of biology, are now obtainable owing to advances in tissue processing and light-sheet microscopy. Recent tissue processing techniques, pioneered by methods like CLARITY \cite{chung2013structural}, allow imaging of whole organs such as the mouse brain at subcellular resolution, by rendering biological tissues transparent. Since then, major improvements have enhanced biomolecular preservation \cite{park2019protection}, structural integrity \cite{ku2020}, and uniform labeling of proteins and nucleic acids across large organs. Combined with volumetric imaging modalities, these advances enable extraction of rich molecular and cellular information from intact tissues \cite{park2024science}. Traditional volumetric imaging methods such as confocal and two-photon microscopy rely on raster scanning, which is inherently slow and depth-limited. Light-sheet fluorescence microscopy (LSFM) instead illuminates entire planes, enabling rapid, low-photodamage imaging of large samples at single-cell resolution. We generated the CANVAS dataset using SmartSPIM, a light-sheet microscope incorporating dynamic axial sweeping \cite{Dean2015-ft}, producing near-isotropic, high-quality datasets spanning the whole mouse brain.

Access to such high-resolution, whole-brain volumetric data offers new opportunities to investigate biological systems in an integrated and spatially continuous manner. In contrast to conventional sub-sampling of tissue by sectioning, intact tissue image data preserves connectivity information and is free from sampling bias that can provide more complete and objective data interpretation. These characteristics are integral to answering key biological questions including those related to neuro-circuitry, spatial heterogeneity, and disease mechanisms across the brain.

Beyond serving as a standalone imaging modality, LSFM fills a crucial gap among existing large-scale biological datasets. Current modalities such as functional magnetic resonance imaging (fMRI), electron microscopy (EM), and transcriptomic profiling provide valuable but fundamentally incomplete perspectives on brain structure and function due to trade-offs in resolution, spatial extent, and molecular specificity. LSFM uniquely bridges these scales by providing subcellular-resolution imaging across entire intact organs, enabling direct visualization of individual cells and their microenvironment.
\vspace{-7pt} 
\paragraph{Functional MRI (fMRI).}
Functional MRI captures large-scale brain activity through hemodynamic measurements, providing whole-brain coverage but at coarse spatial resolution (millimeter-scale voxels containing millions of cells) and limited molecular information. While fMRI enables powerful studies of network-level dynamics and cognition \cite{mri}, it lacks cellular specificity and does not directly measure neuronal activity or morphology. LSFM complements fMRI by providing a single-cell and subcellular structural context for functionally defined pathways and regions through large-scale imaging.
\vspace{-7pt} 
\paragraph{Electron Microscopy (EM).}
Electron microscopy provides nanometer-scale resolution capable of resolving synapses, membranes, and ultrastructural detail. EM excels at microcircuit reconstruction \cite{zheng2018} due to its such high-resolution capability however, it is restricted to extremely small volumes \cite{Kasthuri2015} especially due to slow imaging speed and heavy computational resource requirements, posing a significant challenge in scaling to organ-level analysis. LSFM exhibits the opposite trade-off, where it offers organ-scale coverage with micrometer-scale resolution, making it an ideal mesoscale complement to EM within multiscale connectomics frameworks.

\vspace{-7pt} 
\paragraph{Transcriptomic and spatial profiling.}
Transcriptomic approaches such as single-cell RNA sequencing (scRNA-seq) \cite{Hwang2018}, mRNA-seq \cite{mrnaseq}, and spatial transcriptomics methods \cite{merfish} reveal molecular identity, gene expression, and cell-type diversity across the brain. However, these techniques typically lack full morphological context and do not capture detailed 3D spatial relationships between cells or microenvironmental features. LSFM provides the missing structural dimension by visualizing molecularly labeled cells at subcellular resolution, enabling integrative analyses that couple transcriptomic identity with physical organization and morphology. This integrative potential is especially powerful for cell types such as microglia and astrocytes, whose morphologies are closely linked to their functional states. Immune cells exhibit dynamic structural changes that cannot be inferred from transcriptomics alone, and LSFM makes it possible by enabling a comprehensive morphological profiling at the subcellular level in both molecularly and structurally throughout the brain.

\vspace{-7pt} 
\paragraph{}
Taken together, existing large-scale biomedical data modalities provide valuable perspectives on the brain. Each modality preserves a unique position for specific range of studies and applications, and LSFM sits between these modalities, enabling subcellular-resolution imaging across entire intact organs while preserving molecular labeling and cellular morphology. This complementary role supports applications ranging from brain-wide cell-type mapping and morphological phenotyping to disease pathology, multimodal data integration, and foundational representation learning for large-scale biological imaging. Despite its potential, the field lacks standardized, publicly available whole-brain LSFM datasets with broad cell-type coverage and ground-truth annotations. CANVAS addresses this challenge by providing the first large-scale, multi-marker LSFM benchmark designed for evaluating model generalization and advancing robust foundational models for biological volumetric data.

\section{CANVAS: Whole-brain Volumetric Data}
\label{sec:canvas}
\begin{figure*}[h!]
  \centering
  \begin{subfigure}{0.214\linewidth}
    \includegraphics[width=\linewidth]{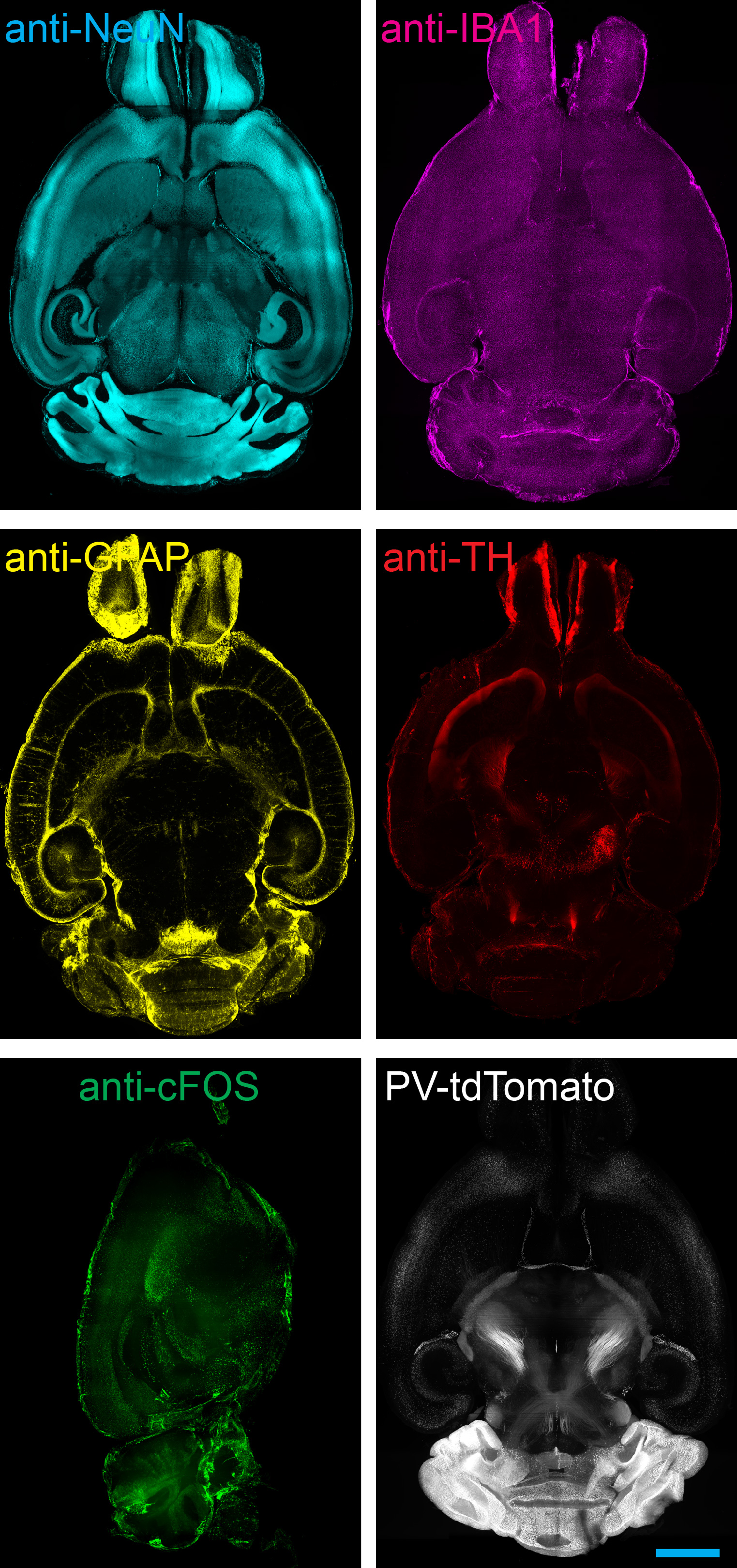}
    \caption{Whole-brain LSFM dataset.}
    \label{fig:datafigure-a}
  \end{subfigure}
  \hfill
  \begin{subfigure}{0.74\linewidth}
    \includegraphics[width=\linewidth]{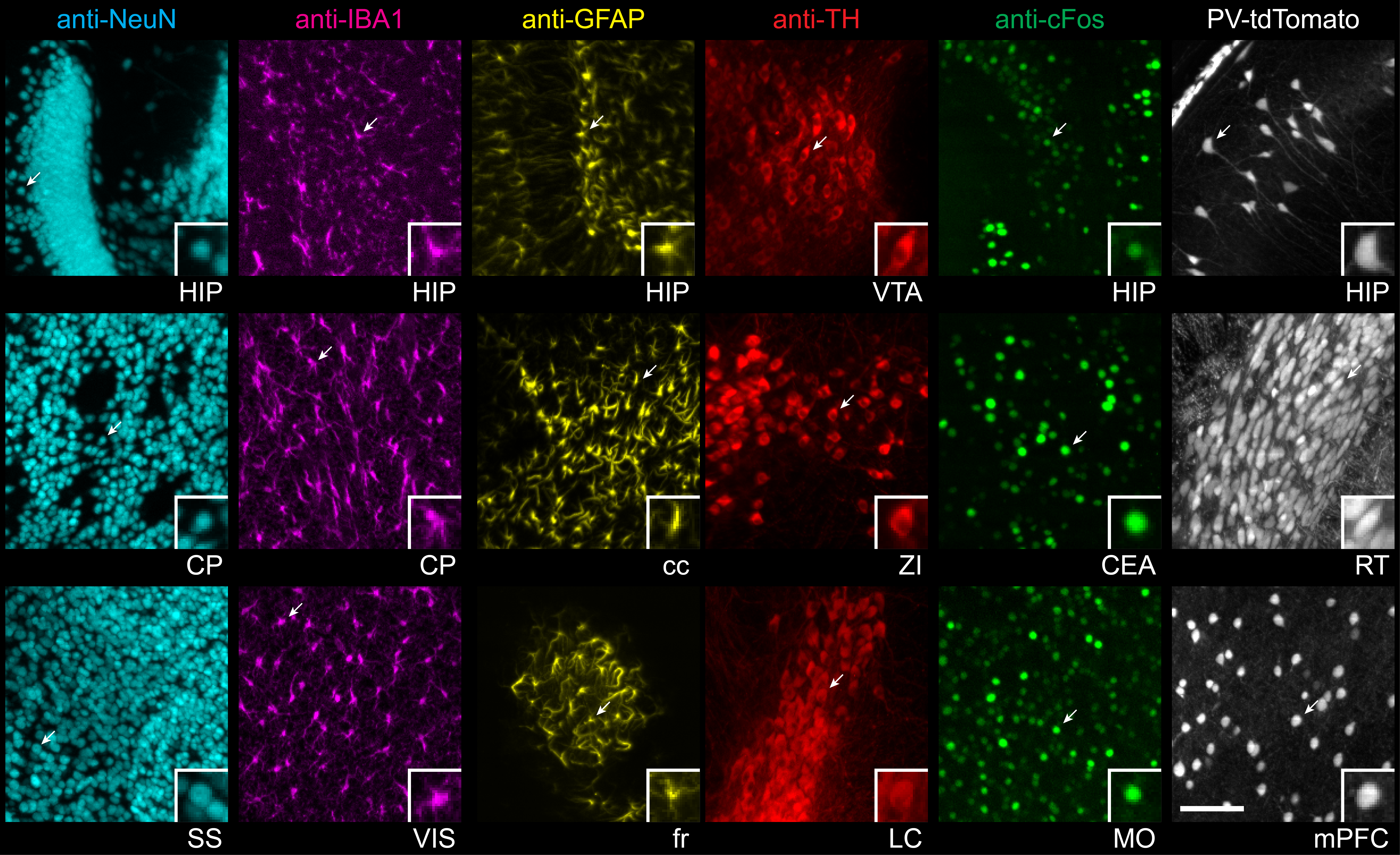}
    \caption{Zoomed-in views of each marker dataset.}
    \label{fig:datafigure-b}
  \end{subfigure}
  \caption{CANVAS Dataset. (a) Overview of datasets in CANVAS showing six cell type markers imaged using light sheet fluorescence microscopy. Markers include NeuN (cyan), IBA1 (magenta), GFAP (yellow), TH (red), cFOS (green), and PV (grey). All markers except PV are based on immunolabeling; PV is transgenically labeled with fluorescent proteins. All datasets represent whole-brain imaging, except cFOS, which is hemisphere-only. Images are 500 µm maximum intensity projections. Scale bar: 2 mm. (b) Zoomed-in views from the CANVAS dataset showing six cell type markers across brain regions, with single-cell insets at bottom right of each panel: NeuN (cyan), IBA1 (magenta), GFAP (yellow), TH (red), cFOS (green), and PV (grey), presented as 80 µm maximum intensity projections. Images include the following brain regions: HIP (hippocampus), VTA (ventral tegmental area), CP (caudate putamen), ZI (zona incerta), CEA (central amygdalar nucleus), RT (reticular nucleus of the thalamus), SS (somatosensory areas), VIS (visual areas), LC (locus coeruleus), MO (somatomotor areas), and mPFC (medial prefrontal cortex); and fiber tracts: cc (corpus callosum) and fr (fasciculus retroflexus). Scale bar: 100 µm. }
  \label{fig:canvas-dataset-image}
\end{figure*}

CANVAS showcases six distinct cell type markers: Neuron Specific Nuclear Protein (NeuN), Ionized calcium-binding adaptor molecule 1 (IBA1), glial fibrillary acidic protein (GFAP), Tyrosine hydroxylase (TH), cFos (a neural activity marker), and Parvalbumin (PV), shown in \Cref{fig:canvas-dataset-image}. Molecular cell type markers are invaluable tools in neuroscience and biomedical research, enabling investigation of cell populations in complex tissues like the brain. The six protein-based markers selected for this benchmark represent only a small subset of available markers, but each highlights a distinct and functionally important cell class that contributes to brain health and disease. Beyond their functional and disease-related significance, these markers exhibit distinct morphological characteristics that vary by brain region.

NeuN is a marker that localizes neuronal cell bodies across the brain. NeuN expression is ubiquitous throughout the brain with varying regional densities. IBA1 labels microglia, resident immune cells, also with ubiquitous expression, with regional differences in both density and morphology that reflect local immune states. GFAP marks astrocytes, which are highly concentrated along the fiber tracts and surrounding vasculature; astrocytes exhibit significant regional and morphological heterogeneity. TH is a classic marker for dopaminergic neurons, found in high densities within several subcortical nuclei, with extensive long-range axonal projections throughout the brain. PV labels one of the largest classes of interneurons, displaying variable densities and morphologies across the cortical and subcortical structures, as well as the cerebellum. cFOS, an immediate early gene product, serves as a proxy for recent neuronal activation, providing a brain-wide snapshot of activity patterns in response to stimuli or behavior. cFOS expression patterns can vary greatly between individual animals but are typically morphologically consistent. Together, these markers encompass a diverse range of cell types and functions, capturing the spatial complexity and regional specialization of the intact brain.
\vspace{-3pt} 

\subsection{Data acquisition}
The workflow for data acquisition begins with SHIELD preservation \cite{park2019protection} of mouse brains. This process helps preserve the integrity of biomolecules during extended tissue processing. Next, delipidation removes lipids, the major component of cellular membranes, to improve molecular and optical access for downstream immunolabeling and subsequent imaging. Fluorescently labeled molecular probes are then uniformly and efficiently delivered throughout intact samples via the SmartBatch+ system, which utilizes probe-binding affinity modulation and electrophoretically enhanced molecular transport \cite{yun2025uniform, kim2015stochastic}. Finally, samples are rendered transparent through refractive-index matching with EasyIndex and imaged using SmartSPIM at 3.6X magnification, generating whole-brain datasets with voxel sizes of 1.8 µm $\times$ 1.8 µm $\times$ 4 µm (x, y, z). Acquisition of a single channel whole brain dataset takes approximately 40 minutes and generates roughly 100 GB of data. Detailed sample preparation steps are described in \Cref{sec:supp_data_acquisition}.

\subsection{Data processing and visualization}
A series of post-processing steps are applied after LSFM data acquisition to improve image quality while preserving underlying scientific information \cite{Swaney2019-vh}. First, destriping is performed using an FFT–wavelet hybrid filter \cite{Munch2009} to remove stripe artifacts introduced by the light paths during SmartSPIM imaging. Filter parameters were tuned across hundreds of samples to suppress stripe artifacts while prioritizing preservation of biological signal; in practice, we observe improved signal consistency without introducing artificial structures or measurable signal degradation. 

Second, because multiple z-stack tiles are required to cover the entire mouse brain, each tile stack must be stitched together, producing a series of TIFF images compressed with lossless zlib (level 1). Each TIFF file represents a complete XY plane at a single z-step determined during imaging (4 µm along the z-axis in our case). Due to the Gaussian profile of the light sheet, tiling artifacts can occur near the seams of stitched tiles. To mitigate this, a channel-specific flat-field correction is computed and applied to every z-plane to reduce large horizontal intensity bands across the dataset. The stitched data are then converted into the Zarr format \cite{zarr2025}, which enables efficient data loading of large volumetric datasets. Finally, the Zarr-formatted data are served via Neuroglancer \cite{neuroglancer2021} for interactive visualization.

\section{Cell Detection Benchmark}
\label{sec:benchmark}
An adult mouse brain contains approximately 70 million neurons and 17 million immune cells. Given the existence of thousands of distinct cell types in the brain, the markers we selected are expressed in hundreds of thousands to millions of cells. For example, microglia, labeled by IBA1, account for 5-12\% of the total cellular population. Thus, detecting individual cells labeled by each marker across the brain requires automated and scalable computational approaches. Accurate cell annotations for LSFM data remain labor-intensive; however, compared to full instance segmentation (\Cref{sec:prelim_eval_seg_perf}), cell centroids offer a more reliable and computationally efficient alternative. Even still, annotated volume requires extensive expert review to verify cell morphology, validate molecular-specific expression profiles, filter false positives, and ensure region-specific consistency. The effort scales poorly given the size and heterogeneity of intact whole-brain datasets. Even though CANVAS provides the largest collection of cell annotations available for whole-brain LSFM, the labeled set still represents only a tiny fraction of the total cellular diversity present in the data. Over the past decade, deep neural networks (DNNs) have achieved state-of-the-art performance in computer vision tasks, including object detection \cite{conf/nips/RenHGS15, DBLP:journals/corr/RedmonDGF15, DBLP:journals/corr/abs-2201-09792, DBLP:journals/corr/abs-2005-12872}. In principle, these models can be applied to localize individual cells in LSFM data. Here, we propose a cell detection benchmark using the CANVAS dataset, along with baseline models built on established DNN backbones. We further show that these baseline models struggle to generalize across anatomical regions, especially when cells exhibit diverse morphological profiles or when training labels are limited.
Lastly, to address the inherent scarcity and high cost of LSFM annotations, we explore self-supervised feature learning using a DINOv2 \cite{oquab2024dinov2learningrobustvisual} Vision Transformer. Such label-free representation learning offers a complementary strategy to supervised detection by enabling the model to learn rich volumetric features directly from large quantities of unlabeled CANVAS data. These ViT-based features provide a promising foundation for building future generalizable models for large-scale LSFM data interpretation.

%%%%%%%%%%%%%%%%%%%%%%%%%%%%%%%%%%%%%%%%%%%%%%%%%%%%%%%%%%%%%%%%%%%%
%  DATA ANNOTATION
%%%%%%%%%%%%%%%%%%%%%%%%%%%%%%%%%%%%%%%%%%%%%%%%%%%%%%%%%%%%%%%%%%%%

\subsection{Data annotation}
For each cell-type marker dataset, three regions of interest (ROIs) were selected for the training set and three for the test set. To generate ground-truth cell annotations, we first performed inference using the baseline models described in Section \ref{baseline_models}. Based on the predicted cell centroids, we center-cropped volumes around each cell with dimensions of 32 $\times$ 32 $\times$ 8 pixels (x, y, z), covering the physical area of 57.6 µm $\times$ 57.6 µm $\times$ 32 µm. Each cropped cell volume was then manually annotated as either $0$ (non-cell) or $1$ (cell). Annotation work was performed by seven annotators using the MorPheT annotation tool \cite{Mappingt95:online}, generating 109--9,577 cell annotations after filtering out volumes labeled as $0$ (non-cells). This approach does not handle false negatives (FNs) missed by the predictions, and these were subsequently recovered through manual review using WebKnossos \cite{boergens2017webknossos}. We first overlaid cell annotations (after filtering out false positives) on the raw data, and manually recovered missing centroids. For marker datasets (TH, GFAP and IBA1) with poor model inference performance, the annotations were completely generated from scratch, using the Webknossos annotation feature. In total, 167,950 predictions were investigated, yielding approximately 93,000 ground-truth cell centroids. \Cref{tab:annotation} shows the details of ROI selection for each cell-type marker dataset for the train set, along with the number of cells annotated. The test set also contains three regions per marker dataset, while not reported here.

\begin{table}[t]
  \centering
  \caption{Regions selected for annotation in each dataset, with per-marker summary statistics (HIP: Hippocampal region, SSP: Primary somatosensory area, CP: Caudoputamen, MOp: Primary motor area, CEA: Central amygdalar nucleus, VTA: Ventral tegmental area, LC: Locus ceruleus, ZI/A13: Dopaminergic A13 group, RT: Reticular nucleus of the thalamus, mPFC: medial Prefrontal cortex, VIS: Visual cortex).}
  \label{tab:annotation}

  \footnotesize
  \setlength{\tabcolsep}{4pt}
  \renewcommand{\arraystretch}{1.05}

  \begin{tabular}{@{} l c c r @{}} 
    \toprule
    \textbf{Marker} & \textbf{ROI} & \textbf{Size (px)} & \textbf{\# Cells} \\
    \midrule

    % ---------------- NeuN ---------------- %
    \multirow{4}{*}{NeuN}
      & HIP & $300\times300\times150$ & 7{,}233 \\
      & CTX(SSp) & $300\times300\times150$ & 1{,}766 \\
      & CP (striatum) & $300\times300\times150$ & 8{,}321 \\
      \cmidrule(lr){2-4}
      & \multicolumn{2}{c}{\textit{Summary}} 
      & \textit{17{,}320,\; 5{,}773 $\pm$ 2{,}868} \\

    \midrule
    % ---------------- cFos ---------------- %
    \multirow{4}{*}{cFos}
      & HIP & $500\times500\times250$ & 3{,}782 \\
      & CTX(MOp) & $300\times300\times150$ & 2{,}827 \\
      & CEA & $400\times400\times200$ & 2{,}781 \\
      \cmidrule(lr){2-4}
      & \multicolumn{2}{c}{\textit{Summary}} 
      & \textit{9{,}390,\; 3{,}130 $\pm$ 461} \\

    \midrule
    % ---------------- TH ---------------- %
    \multirow{4}{*}{TH}
      & VTA & $300\times300\times150$ & 1{,}178 \\
      & LC & $300\times300\times150$ & 1{,}118 \\
      & Z1/A13 & $300\times300\times150$ & 643 \\
      \cmidrule(lr){2-4}
      & \multicolumn{2}{c}{\textit{Summary}} 
      & \textit{2{,}939,\; 980 $\pm$ 239} \\

    \midrule
    % ---------------- PV ---------------- %
    \multirow{4}{*}{PV}
      & HIP & $600\times600\times300$ & 949 \\
      & RT & $400\times400\times200$ & 4{,}904 \\
      & CTX(mPFC) & $500\times500\times250$ & 1{,}543 \\
      \cmidrule(lr){2-4}
      & \multicolumn{2}{c}{\textit{Summary}} 
      & \textit{7{,}396,\; 2{,}465 $\pm$ 1{,}741} \\

    \midrule
    % ---------------- GFAP ---------------- %
    \multirow{4}{*}{GFAP}
      & HIP & $300\times300\times150$ & 1{,}944 \\
      & fiber tracts (cc) & $200\times200\times100$ & 477 \\
      & fiber tracts (fr) & $300\times300\times150$ & 109 \\
      \cmidrule(lr){2-4}
      & \multicolumn{2}{c}{\textit{Summary}} 
      & \textit{2{,}530,\; 843 $\pm$ 793} \\

    \midrule
    % ---------------- IBA1 ---------------- %
    \multirow{4}{*}{IBA1}
      & HIP & $400\times400\times200$ & 2{,}076 \\
      & VIS1-5 & $400\times400\times200$ & 2{,}021 \\
      & CP & $400\times400\times200$ & 2{,}073 \\
      \cmidrule(lr){2-4}
      & \multicolumn{2}{c}{\textit{Summary}} 
      & \textit{6{,}170,\; 2{,}057 $\pm$ 25} \\

     \bottomrule
  \end{tabular}
\end{table}

\subsection{Baseline models}
\label{baseline_models}
U-Net \cite{DBLP:journals/corr/RonnebergerFB15} and ResNet \cite{DBLP:journals/corr/HeZRS15} architectures are among the most widely used backbones in computer vision tasks for biomedical image analysis. Numerous variations of these networks have been proposed \cite{DBLP:journals/corr/abs-1806-05034}, many incorporating 3D convolution layers to handle volumetric data \cite{DBLP:journals/corr/CicekALBR16,Cai2023-zw}. More recently, models based on the Vision Transformer (ViT) architecture \cite{DBLP:journals/corr/abs-2010-11929} have also been introduced. However, most of these existing models were developed and trained for other data modalities, such as computed tomography (CT), functional magnetic resonance imaging (fMRI), X-ray, or histopathology \cite{HAO2024105365, 10178853, Sheng2023-xj, WANG2023104976}, with relatively few tailored for microscopy data---and even then, typically for specific subdomains (e.g., retina confocal imaging) \cite{Chen2022-cj}. In addition, training ViT models requires extensive computational resources, even for the smallest networks, and they are generally inefficient in terms of inference speed when applied to large-scale datasets such as CANVAS. For our benchmark, we adopted ConvMixer \cite{DBLP:journals/corr/abs-2201-09792} as the backbone for baseline models. ConvMixer employs a concept similar to patch embeddings in ViT but is simpler and sufficiently effective on our dataset to serve as a baseline. As ConvMixer was originally proposed for semantic segmentation, we introduced an additional layer, called \textbf{FindMaxima layer}, to transform the network's probability heatmap into discrete 3D cell locations using non-maximum suppression (\Cref{algo1}). With this layer, a voxel at position $(x, y, z)$ is identified as a local maximum and defined as a cell centroid if:
\vspace{-2pt} 
\begin{equation}
H(x, y, z) = \max_{\substack{|i| \leq d_{min} \\ |j| \leq d_{min} \\ |k| \leq d_{min}}} H(x+i, y+j, z+k)
\end{equation}
where $H$ is the 3D probability heatmap derived from the model, $d_{min}$ is the minimum distance between peaks, and a valid detection satisfies:
\vspace{-2pt} 
\begin{align}
\text{Valid}(x, y, z)
    &= [H(x,y,z) = H_{\max}(x,y,z)] \nonumber\\
    &\quad \land [H(x,y,z) \ge \tau]
\label{eq2}
\end{align}
where $H_{max}$ represents 3D max-pooled output from $H$ for finding local maxima, and $\tau$ is a threshold ranged from 0 to 1 (\Cref{alg:algo2}).

\BlankLine
\vspace{-4pt}
\begin{algorithm}[htbp]
\caption{FindMaxima Layer}
\label{algo1}
\footnotesize
\SetAlgoLined
\SetKwInOut{Input}{Input}
\SetKwInOut{Output}{Output}
\SetKwInOut{Parameters}{Parameters}
\SetKw{Return}{return}
\SetKwFunction{Where}{where}
\SetKwFunction{Cast}{cast}
\SetKwFunction{Min}{min}
\SetKwFunction{ExpandDims}{expand\_dims}
\SetKwFunction{Concat}{concatenate}
\Input{Heatmap $H$ (shape $(B, X, Y, Z)$), max-pooled heatmap $H_{max}$ (shape $(B, X, Y, Z)$), batch indices $I$ (shape $(B,)$)}
\Output{Cell locations $L$ (shape $(N, 4)$) with columns $[b, x, y, z]$}
\Parameters{Threshold $\tau \in [0,1]$}

\BlankLine
    $M_{local} \gets (H = H_{max})$
    
    $M_{thresh} \gets (H \geq \tau)$
    
    $M_{valid} \gets M_{local} \land M_{thresh}$

    $L \gets \Where(M_{valid})$ 

    $i_{min} \gets \Cast(\Min(I), \text{int64})$

    $L_{batch} \gets L[:, 0] + i_{min}$

    $L_{batch} \gets \ExpandDims(L_{batch}, \text{axis}=-1) $

    $L \gets \Concat([L_{batch}, L[\cdot, 1:]], \text{axis} = 1)$

    \Return $L$
\end{algorithm}

\BlankLine
\vspace{-4pt}
To train the network, we manually generated a small size of cell mask training data for each cell-type marker, using an additional whole-brain dataset not included in this paper. We limited the size of the training set to the minimum sufficient to guide cell centroid annotations due to the impracticality of generating whole-brain cell masks. A binary form of focal loss \cite{DBLP:journals/corr/abs-1708-02002} was used for model training and a separate model was trained for each cell-type marker dataset to further investigate their generalizability. An NVIDIA RTX 3090 or 4090 graphics card was used for training, and the model converged within one day.

\subsubsection{Evaluation method}
\label{evaluation}
Model performance was evaluated using accuracy and the F1 score. True positives (TPs) were determined by calculating the Euclidean distance between each predicted cell coordinate and the closest ground-truth cell center using a kd-tree nearest-neighbor search \cite{bentley1975multidimensional}. A prediction was assigned as a TP if the distance fell below a tolerance threshold, defined in voxels and set individually for each cell type, shown in \Cref{table_tolerance}. The tolerance threshold was set to the average cell radius, estimated as half the mean diameter of six representative cells sampled from each anatomical region. Each prediction was matched to at most one ground-truth cell center in a greedy fashion to enforce a one-to-one correspondence. False positives (FPs) and false negatives (FN) were then calculated using the remaining unmatched predictions and ground-truth centers. Accuracy was calculated as $\frac{TP}{TP+FP+FN}$, precision was calculated as $\frac{TP}{TP+FP}$, and recall was calculated as $\frac{TP}{TP+FN}$. F1 score was calculated as the harmonic mean of precision and recall. This evaluation method will also be applied to the models submitted to the leaderboard.

\begin{table}[t]
\centering
\caption{Tolerance threshold (in voxels) per dataset. Each threshold was applied across all regions per marker dataset.}
\label{table_tolerance}
\footnotesize
\setlength{\tabcolsep}{8pt}
\renewcommand{\arraystretch}{1.05}
\begin{tabular}{@{} l c c c c c c @{}} 
\toprule
\textbf{Marker} & NeuN & cFos & TH & PV & GFAP & IBA1 \\
\midrule
\textbf{Tolerance} & 6.0 & 6.0 & 8.0 & 8.0 & 8.0 & 5.0 \\
\bottomrule
\end{tabular}
\end{table}

\begin{table*}[t]
\centering
\caption{Cross-dataset performance of each model across cell-type datasets. (\textbf{Bold} indicates evaluation on matched data; \underline{underline} indicates the highest-performing model.)}
\label{tab:cross_dataset_summary}
\small
\setlength{\tabcolsep}{5pt}
\renewcommand{\arraystretch}{1.15}
\begin{tabular}{lcccccccccccc}
\toprule
& \multicolumn{2}{c}{cFos Model}
& \multicolumn{2}{c}{NeuN Model}
& \multicolumn{2}{c}{TH Model}
& \multicolumn{2}{c}{PV Model}
& \multicolumn{2}{c}{GFAP Model}
& \multicolumn{2}{c}{Iba1 Model} \\
\cmidrule(lr){2-3}
\cmidrule(lr){4-5}
\cmidrule(lr){6-7}
\cmidrule(lr){8-9}
\cmidrule(lr){10-11}
\cmidrule(lr){12-13}
\textbf{Model}
& \textbf{Acc} & \textbf{F1}
& \textbf{Acc} & \textbf{F1}
& \textbf{Acc} & \textbf{F1}
& \textbf{Acc} & \textbf{F1}
& \textbf{Acc} & \textbf{F1}
& \textbf{Acc} & \textbf{F1} \\
\midrule
cFos & \underline{\textbf{0.64}} & \underline{\textbf{0.78}} & 0.62 & 0.76 & 0.15 & 0.26 & 0.59 & 0.74 & 0.17 & 0.29 & 0.58 & 0.74 \\
NeuN & 0.01 & 0.02 & \underline{\textbf{0.67}} & \underline{\textbf{0.81}} & 0.12 & 0.21 & 0.57 & 0.73 & 0.02 & 0.04 & 0.40 & 0.57 \\
TH & 0.02 & 0.04 & 0.25 & 0.41 & \underline{\textbf{0.40}} & \underline{\textbf{0.57}} & 0.11 & 0.20 & 0.08 & 0.14 & 0.16 & 0.28 \\
PV & 0.16 & 0.28 & \underline{0.81} & \underline{0.89} & 0.52 & 0.68 & \textbf{0.46} & \textbf{0.63} & 0.12 & 0.21 & 0.45 & 0.62 \\
GFAP & 0.00 & 0.00 & 0.03 & 0.05 & 0.02 & 0.04 & 0.00 & 0.01 & \textbf{0.20} & \textbf{0.33} & \underline{0.44} & \underline{0.61} \\
Iba1 & 0.01 & 0.03 & 0.27 & 0.43 & 0.39 & 0.56 & 0.03 & 0.06 & 0.28 & 0.43 & \underline{\textbf{0.69}} & \underline{\textbf{0.81}} \\
\bottomrule
\end{tabular}
\end{table*}

\subsubsection{Results}
\label{results}
Using the baseline models and annotations described above, we evaluated model performance both within datasets and across datasets using six models, each trained on a different cell-type marker. Across most datasets, each model performed best on its corresponding marker dataset, indicating strong dataset-specific specialization but limited cross-dataset generalization, with GFAP as an exception. Performance also varied across regions, likely due to distinct cell morphologies and region-specific characteristics.
For example, the TH model peaked at an F1 score of 0.31 on the cFos dataset, while reaching 0.83 in region\_6 of the TH dataset, highlighting the challenge the model faces in generalizing across markers and regions. This trend was consistent across most marker and model combinations, as shown in Table \ref{tab:cross_dataset_summary} and \ref{tab:model_results}.
Performance on the PV dataset varied substantially across models, with average F1 scores (across regions) of 0.28 (cFos model), 0.89 (NeuN model), 0.68 (TH model), 0.63 (PV model), 0.21 (GFAP model), and 0.62 (IBA1 model). Inter-regional variation was also observed, with F1 scores ranging from 0.37 (region\_2) to 0.98 (region\_3, region\_6), reflecting region-dependent PV\textsuperscript{+} staining profiles. In the RT (region\_2), the NeuN model performed better, likely due to greater morphological similarity to nuclear signals than to PV\textsuperscript{+} cells in other regions like HIP and mPFC. Further highlighting limited generalization, the TH model showed large regional variation on its own dataset (F1: 0.45--0.83). Lastly, GFAP cell detection remained particularly challenging, with near-zero performance across most models. Even the best results were modest (0.44 for GFAP, 0.67 for IBA1), with the IBA1 model unexpectedly outperforming the GFAP model.

%%%%%%%%%%%%%%%%%%%%%%%%%%%%%%%%%%%%%%%%%%
%% Representation learning
%%%%%%%%%%%%%%%%%%%%%%%%%%%%%%%%%%%%%%%%%%
\subsection{Representation learning with CANVAS}
Although supervised models such as ConvMixer establish a strong baseline for cell detection, the broader challenge in LSFM-based analysis is the unavoidable scarcity of annotated data. Even with CANVAS providing the largest set of cell-level labels available for intact whole-brain LSFM, the annotated volume still remains small relative to the diverse cellular structures, morphologies, and cytoarchitectural contexts present across the brain. The cost and expertise required for further annotation make the scalable supervised approaches challenging. Thus, label-free approaches such as self-supervised learning, particularly transformer-based methods such as DINOv2, which can exploit the vast amount of unlabeled LSFM data becomes more compelling. LSFM datasets contain highly structured spatial information that provide rich supervisory signals under self-distillation or masked-token prediction. These properties make LSFM ideally suited for learning foundational volumetric representations via self-supervised learning.
%\vspace{-7pt} 

\subsubsection{3D-MAE model adaptation}
To explore the feasibility of this direction, we adopted a DINOv2-style ViT architecture with Masked Autoencoder (MAE) \cite{DBLP:journals/corr/abs-2111-06377} pretraining for learning robust representations directly from unlabeled CANVAS volumes. By reconstructing masked 3D cell patches, the ViT encoder learns rich volumetric features without supervision, capturing detailed cellular contexts. The goal is not yet full cell detection, but rather the development of a general-purpose 3D feature extractor capable of encoding cellular and mesoscale anatomical structure, which can offer a strong complementary foundation to our supervised ConvMixer baselines, and later support downstream tasks.

We made two key modifications to adopt the MAE approach. First, we optimized crop and patch sizes for cell morphology which can cover the cell types that CANVAS dataset provides. Given typical dimensions of the neuronal soma (5-15 $\mu$m) and immune cells like microglia and astrocytes (20-40 $\mu$m), we evaluated three configurations of crop/patch pairs that balance spatial resolution and computational efficiency ([16$\times$32$\times$32 / 4$\times$8$\times$8, 24$\times$48$\times$48 / 6$\times$12$\times$12, 32$\times$64$\times$64 / 8$\times$16$\times$16]), maintaining a fixed 4$\times$4$\times$4 patch grid in ($z$, $y$, $x$) to balance spatial resolution and computational efficiency.

Second, we introduced content-aware reconstruction weighting to address the sparsity of biological signals. Standard MAE applies uniform loss weighting, but CANVAS data volumes are highly sparse and exhibit heterogeneous sparsity throughout the brain. We weighted masked patches by their intensity variance:

\begin{equation}
\begin{aligned}
\mathcal{L}_{\text{MAE}} &= \frac{1}{|\mathcal{I}_{\text{mask}}|}
    \sum_{i \in \mathcal{I}_{\text{mask}}} w_i
    \cdot \|\hat{\mathbf{x}}_i - \mathbf{x}_i\|^2, \\[4pt]
w_i &= \alpha + \gamma \cdot \min\!\left(1,\;
    \frac{\operatorname{Var}(\mathbf{x}_i)}{\bar{\sigma}^2}\right)
\end{aligned}
\end{equation}

where background patches receive $w_i \approx \alpha$ ($\alpha = 1$) and cell-containing patches receive $w_i \approx (\alpha + \gamma)$ ($\gamma=9$), prioritizing informative regions without ignoring spatial context.

\subsubsection{Training and hyperparameter optimization}

\begin{figure}[t]
\footnotesize
\centering

% ---- Row 1 ----
\includegraphics[width=\linewidth]{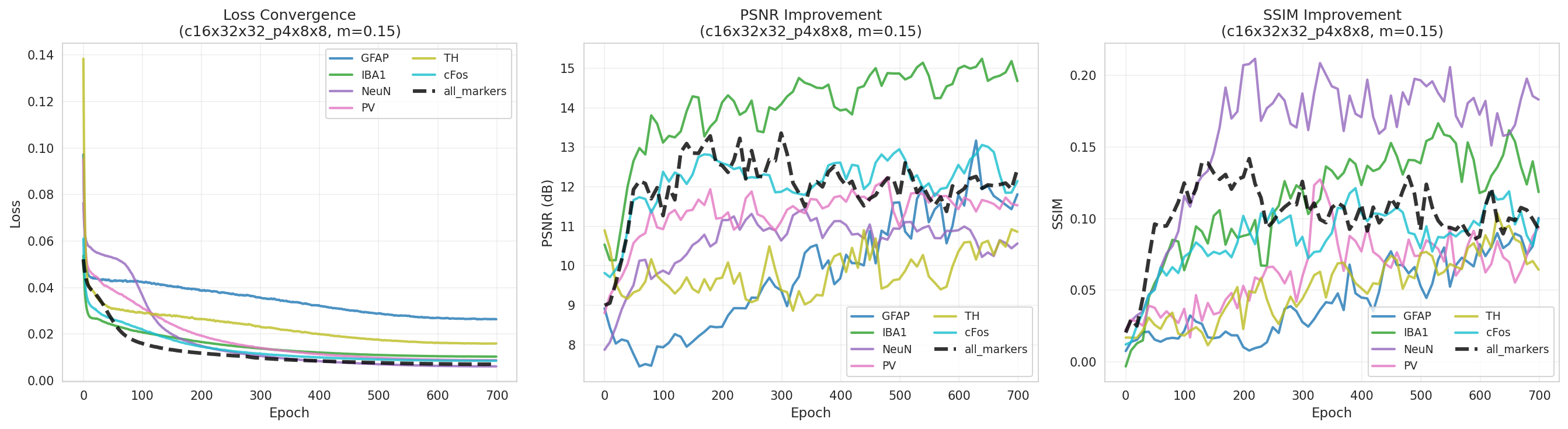}

\vspace{0.3em}

% ---- Row 2 ----
\includegraphics[width=\linewidth]{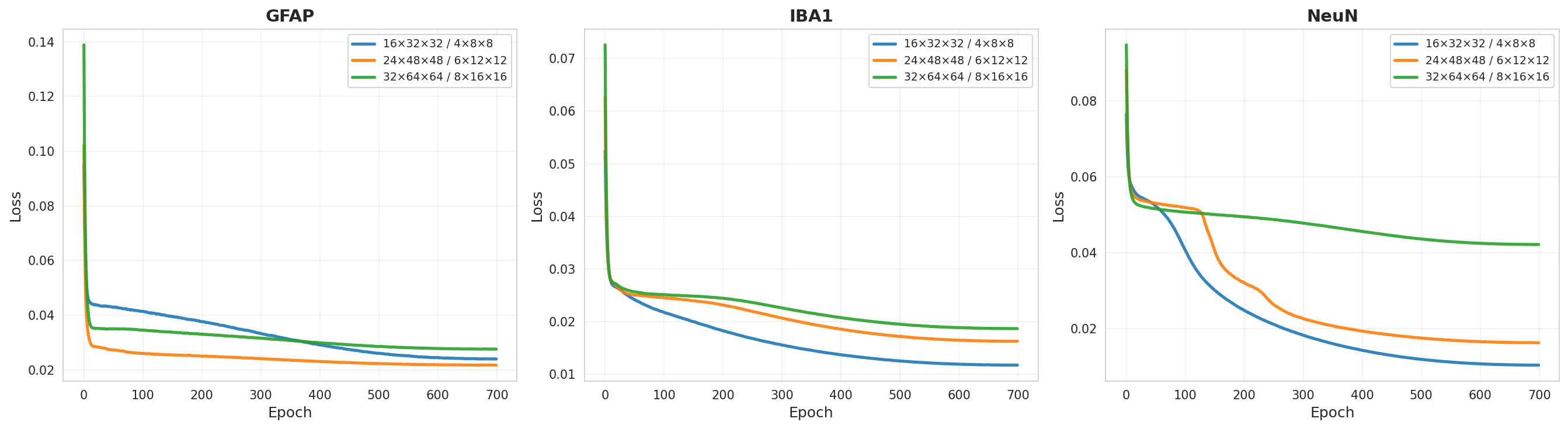}

\caption{3D-MAE training results. \textit{Top:} Convergence (loss, PSNR, SSIM) comparison across all markers plus the all-markers model (dashed lines) using the best configuration (16$\times$32$\times$32/4$\times$8$\times$8, $m=0.15$). \textit{Bottom:} Effect of crop/patch size. GFAP dataset shows the best performance on (24$\times$48$\times$48/6$\times$12$\times$12), while IBA1 and NeuN datasets show the smallest configuration (blue) outperforms larger alternatives.}

\label{fig:mae_training}
\end{figure}

We conducted a systematic Weights \& Biases \cite{wandb} sweep across 84 experiments: 3 crop/patch pairs $\times$ 4 mask ratios $m \in \{0.15, 0.35, 0.55, 0.75\}$ $\times$ 7 datasets (6 individual markers + combined). Unlike natural images where MAE uses $m=0.75$ \cite{DBLP:journals/corr/abs-2111-06377}, we found 3D microscopy images benefit from lower masking, with $m=0.15$ optimal for four datasets with smaller cell bodies. This reflects the higher semantic density of small 3D patches, where masking 75\% removes too much spatial context for accurate reconstruction. The smallest configuration (16$\times$32$\times$32 crop, 4$\times$8$\times$8 patch) achieved the best performance in \textit{all} markers except the GFAP dataset (Fig \ref{fig:mae_training}, Table \ref{tab:mae_results}), likely because:
(1) tight crops reduce the confounding background and
(2) each patch captures a meaningful substructure.

Models were trained for 700 epochs using AdamW ($\eta=1.5 \times 10^{-4}$, weight decay: 0.05, batch size: 64) with cosine annealing to $0.01\eta$. Training curves (Fig.~\ref{fig:mae_training}) show rapid convergence (50\% loss reduction in 100 epochs) and stable optimization without overfitting. We also trained an all-markers model combining $\sim$60k patches from all 6 markers (10$\times$ more data than single-marker models). All-markers models converged competitively with marker-specific models, demonstrating effective knowledge sharing across markers despite morphological diversity. They also achieved competitive reconstruction loss (0.0070 vs. 0.0061 for best single-marker) while providing transferable features for any marker, essential for detecting novel cell types without marker-specific retraining.

\subsubsection{Results}
Our systematic evaluation of 84 3D-MAE configurations revealed three key findings. First, the smallest architecture (4$\times$8$\times$8 patch) consistently outperformed larger alternatives, suggesting tight cell-centered crops reduce confounding signals more than larger receptive fields improve context. Second, optimal mask ratios (0.15-0.55) were substantially lower than natural image MAE (0.75), reflecting the semantic density of 3D cellular patches where excessive masking destroys essential spatial relationships. Third, the all-markers model achieved reconstruction loss within 15\% of best single-marker performance (0.0070 vs. 0.0061 for NeuN) despite training on 10$\times$ more heterogeneous data, demonstrating robust transfer learning across morphologically distinct cell types. Training curves showed rapid convergence, with PSNR (12-16 dB) and SSIM (0.17-0.36) values consistent with noisy microscopy reconstruction.

%-------------------------------------------------------------------------
\subsubsection {Applications}
To examine applicability of the representation learning to the baseline models reported in \ref{baseline_models}, we performed additional experiments using the all-markers 3D-MAE model ([16\texttimes32\texttimes32 / 4\texttimes8\texttimes8], m=0.15). We extracted 384 encoder features and concatenated them with baseline detector features (384 features for NeuN, 512 features for GFAP, after dataset-specific adaptive pooling), then trained a binary classifier to distinguish TPs from FPs among baseline detection candidates. We tested detector-only, 3D-MAE-only, and concatenated feature variants with varying contribution ratios ($r$). Results showed that MAE features can effectively serve as a post-hoc refinement stage, with the greatest impact on cell types where the detector produces many false positives. GFAP showed substantial improvement for all three regions (+22.9\% mean F1, up to +86.3\% on region 3 with 3D-MAE-only), where the baseline model struggled due to complex astrocyte morphology. NeuN showed a modest improvement in region\_2 (+3.0\% with $r=0.3$ \textit{(154 $+$ 384 features)}). These results show that MAE representations effectively separate TPs from FPs by capturing complementary information beyond what the existing classifier provides, highlighting the value of self-supervised learning on vast amounts of unlabeled LSFM data for morphologically challenging cell types.

\section{Conclusion}
\label{conclusion}
In this paper, we present CANVAS, a large-scale and subcellular-resolution LSFM dataset with six different markers and accompanied by extensive annotations. A public leaderboard enables benchmarking of baseline models to advance robust foundational models for the LSFM domain. Current limitations include the limited number of manually generated cell-centroid labels, which future expansions and active-learning strategies can address. Beyond supervised detection, our experiments also show that CANVAS is also well suited for self-supervised representation learning. By adapting a 3D Masked Autoencoder (MAE) with content-aware weighting and morphology-aligned patching, we demonstrate that high-quality volumetric features can be learned directly from unlabeled data, providing a general-purpose encoder that complements supervised baselines and supports downstream tasks with limited annotations. Overall, CANVAS dataset offers an immediately valuable resource for large-scale LSFM analysis and serves as a complementary axis for multimodal integration with fMRI, EM, and mRNA sequencing. Future work will expand marker coverage, increase annotated samples, and further develop self-supervised models for robust, scalable volumetric representation learning.
{
    \small
    \bibliographystyle{ieeenat_fullname}
    \bibliography{main}

@String(CVPR= {IEEE Conf. Comput. Vis. Pattern Recog.})

@String(NIPS= {Adv. Neural Inform. Process. Syst.})

@String(CVPR  = {CVPR})

@String(NIPS  = {NeurIPS})

@ARTICLE{Das2023-yk,
  title    = "Region-specific heterogeneity in neuronal nuclear morphology in
              young, aged and in Alzheimer's disease mouse brains",
  author   = "Das, Soumen and Ramanan, Narendrakumar",
  journal  = "Front. Cell Dev. Biol.",
  volume   =  11,
  pages    = "1032504",
  month    =  feb,
  year     =  2023,
  language = "en"
}

@ARTICLE{Miller_2018, title={Astrocyte heterogeneity in the Adult Central Nervous System}, url={https://pmc.ncbi.nlm.nih.gov/articles/PMC6262303/}, journal={Frontiers in cellular neuroscience}, publisher={U.S. National Library of Medicine}, author={Miller, Sean J}, year={2018}, month={Nov}}

@ARTICLE{Bryant2023-wn,
  title     = "Endothelial cells are heterogeneous in different brain regions
               and are dramatically altered in Alzheimer's disease",
  author    = "Bryant, Annie and Li, Zhaozhi and Jayakumar, Rojashree and
               Serrano-Pozo, Alberto and Woost, Benjamin and Hu, Miwei and
               Woodbury, Maya E and Wachter, Astrid and Lin, Gen and Kwon,
               Taekyung and Talanian, Robert V and Biber, Knut and Karran, Eric
               H and Hyman, Bradley T and Das, Sudeshna and Bennett, Rachel E",
  journal   = "J. Neurosci.",
  publisher = "Society for Neuroscience",
  volume    =  43,
  number    =  24,
  pages     = "4541--4557",
  month     =  jun,
  year      =  2023,
  copyright = "https://creativecommons.org/licenses/by-nc-sa/4.0/",
  language  = "en"
}

@ARTICLE{Ueda2020-tk,
  title     = "Tissue clearing and its applications in neuroscience",
  author    = "Ueda, Hiroki R and Ert{\"u}rk, Ali and Chung, Kwanghun and
               Gradinaru, Viviana and Ch{\'e}dotal, Alain and Tomancak, Pavel
               and Keller, Philipp J",
  journal   = "Nat. Rev. Neurosci.",
  publisher = "Springer Science and Business Media LLC",
  volume    =  21,
  number    =  2,
  pages     = "61--79",
  month     =  feb,
  year      =  2020,
  language  = "en"
}

@article{park2019protection,
  title={Protection of tissue physicochemical properties using polyfunctional crosslinkers},
  author={Park, Young-Gyun and Sohn, Chang Ho and Chen, Ritchie and McCue, Margaret and Yun, Dae Hee and Drummond, Gabrielle T and Ku, Taeyun and Evans, Nicholas B and Oak, Hayeon Caitlyn and Trieu, Wendy and others},
  journal={Nature biotechnology},
  volume={37},
  number={1},
  pages={73--83},
  year={2019},
  publisher={Nature Publishing Group US New York}
}

@article{yun2025uniform,
  title={Uniform volumetric single-cell processing for organ-scale molecular phenotyping},
  author={Yun, Dae Hee and Park, Young-Gyun and Cho, Jae Hun and Kamentsky, Lee and Evans, Nicholas B and DiNapoli, Nicholas and Xie, Katherine and Choi, Seo Woo and Albanese, Alexandre and Tian, Yuxuan and others},
  journal={Nature Biotechnology},
  pages={1--12},
  year={2025},
  publisher={Nature Publishing Group US New York}
}

@article{kim2015stochastic,
  title={Stochastic electrotransport selectively enhances the transport of highly electromobile molecules},
  author={Kim, Sung-Yon and Cho, Jae Hun and Murray, Evan and Bakh, Naveed and Choi, Heejin and Ohn, Kimberly and Ruelas, Luzdary and Hubbert, Austin and McCue, Meg and Vassallo, Sara L and others},
  journal={Proceedings of the National Academy of Sciences},
  volume={112},
  number={46},
  pages={E6274--E6283},
  year={2015},
  publisher={National Academy of Sciences}
}

@article{chung2013structural,
  title={Structural and molecular interrogation of intact biological systems},
  author={Chung, Kwanghun and Wallace, Jenelle and Kim, Sung-Yon and Kalyanasundaram, Sandhiya and Andalman, Aaron S and Davidson, Thomas J and Mirzabekov, Julie J and Zalocusky, Kelly A and Mattis, Joanna and Denisin, Aleksandra K and others},
  journal={Nature},
  volume={497},
  number={7449},
  pages={332--337},
  year={2013},
  publisher={Nature Publishing Group UK London}
}

@article{DBLP:journals/corr/HeZRS15,
  author       = {Kaiming He and
                  Xiangyu Zhang and
                  Shaoqing Ren and
                  Jian Sun},
  title        = {Deep Residual Learning for Image Recognition},
  journal      = {CoRR},
  volume       = {abs/1512.03385},
  year         = {2015},
  url          = {http://arxiv.org/abs/1512.03385},
  eprinttype    = {arXiv},
  eprint       = {1512.03385},
  bibsource    = {dblp computer science bibliography, https://dblp.org}
}

@article{DBLP:journals/corr/RonnebergerFB15,
  author       = {Olaf Ronneberger and
                  Philipp Fischer and
                  Thomas Brox},
  title        = {U-Net: Convolutional Networks for Biomedical Image Segmentation},
  journal      = {CoRR},
  volume       = {abs/1505.04597},
  year         = {2015},
  url          = {http://arxiv.org/abs/1505.04597},
  eprinttype    = {arXiv},
  eprint       = {1505.04597},
  bibsource    = {dblp computer science bibliography, https://dblp.org}
}

@article{DBLP:journals/corr/CicekALBR16,
  author       = {{\"{O}}zg{\"{u}}n {\c{C}}i{\c{c}}ek and
                  Ahmed Abdulkadir and
                  Soeren S. Lienkamp and
                  Thomas Brox and
                  Olaf Ronneberger},
  title        = {3D U-Net: Learning Dense Volumetric Segmentation from Sparse Annotation},
  journal      = {CoRR},
  volume       = {abs/1606.06650},
  year         = {2016},
  url          = {http://arxiv.org/abs/1606.06650},
  eprinttype    = {arXiv},
  eprint       = {1606.06650},
  bibsource    = {dblp computer science bibliography, https://dblp.org}
}

@ARTICLE{Cai2023-zw,
  title    = "Swin {Unet3D}: a three-dimensional medical image segmentation
              network combining vision transformer and convolution",
  author   = "Cai, Yimin and Long, Yuqing and Han, Zhenggong and Liu, Mingkun
              and Zheng, Yuchen and Yang, Wei and Chen, Liming",
  journal  = "BMC Med. Inform. Decis. Mak.",
  volume   =  23,
  number   =  1,
  pages    = "33",
  month    =  feb,
  year     =  2023,
  language = "en"
}

@article{DBLP:journals/corr/abs-1806-05034,
  author       = {Simon A. A. Kohl and
                  Bernardino Romera{-}Paredes and
                  Clemens Meyer and
                  Jeffrey De Fauw and
                  Joseph R. Ledsam and
                  Klaus H. Maier{-}Hein and
                  S. M. Ali Eslami and
                  Danilo Jimenez Rezende and
                  Olaf Ronneberger},
  title        = {A Probabilistic U-Net for Segmentation of Ambiguous Images},
  journal      = {CoRR},
  volume       = {abs/1806.05034},
  year         = {2018},
  url          = {http://arxiv.org/abs/1806.05034},
  eprinttype    = {arXiv},
  eprint       = {1806.05034},
  bibsource    = {dblp computer science bibliography, https://dblp.org}
}

@article{DBLP:journals/corr/abs-2010-11929,
  author       = {Alexey Dosovitskiy and
                  Lucas Beyer and
                  Alexander Kolesnikov and
                  Dirk Weissenborn and
                  Xiaohua Zhai and
                  Thomas Unterthiner and
                  Mostafa Dehghani and
                  Matthias Minderer and
                  Georg Heigold and
                  Sylvain Gelly and
                  Jakob Uszkoreit and
                  Neil Houlsby},
  title        = {An Image is Worth 16x16 Words: Transformers for Image Recognition
                  at Scale},
  journal      = {CoRR},
  volume       = {abs/2010.11929},
  year         = {2020},
  url          = {https://arxiv.org/abs/2010.11929},
  eprinttype    = {arXiv},
  eprint       = {2010.11929},
  bibsource    = {dblp computer science bibliography, https://dblp.org}
}

@article{HAO2024105365,
title = {DBM-ViT: A multiscale features fusion algorithm for health status detection in CXR / CT lungs images},
journal = {Biomedical Signal Processing and Control},
volume = {87},
pages = {105365},
year = {2024},
issn = {1746-8094},
doi = {https://doi.org/10.1016/j.bspc.2023.105365},
url = {https://www.sciencedirect.com/science/article/pii/S174680942300798X},
author = {Yong Hao and Chengxiang Zhang and Xiyan Li}
}

@INPROCEEDINGS{10178853,
  author={El Badaoui, Rim and Coll, Ester Bonmati and Psarrou, Aleka and Villarini, Barbara},
  booktitle={2023 IEEE 36th International Symposium on Computer-Based Medical Systems (CBMS)}, 
  title={3D CATBraTS: Channel Attention Transformer for Brain Tumour Semantic Segmentation}, 
  year={2023},
  volume={},
  number={},
  pages={489-494},
  doi={10.1109/CBMS58004.2023.00267}}

@ARTICLE{Sheng2023-xj,
  title     = "Transformer-based deep learning network for tooth segmentation
               on panoramic radiographs",
  author    = "Sheng, Chen and Wang, Lin and Huang, Zhenhuan and Wang, Tian and
               Guo, Yalin and Hou, Wenjie and Xu, Laiqing and Wang, Jiazhu and
               Yan, Xue",
  journal   = "J. Syst. Sci. Complex.",
  publisher = "Springer Science and Business Media LLC",
  volume    =  36,
  number    =  1,
  pages     = "257--272",
  year      =  2023,
  copyright = "https://www.springernature.com/gp/researchers/text-and-data-mining",
  language  = "en"
}

@ARTICLE{Chen2022-cj,
  title     = "{PCAT-UNet}: {UNet-like} network fused convolution and
               transformer for retinal vessel segmentation",
  author    = "Chen, Danny and Yang, Wenzhong and Wang, Liejun and Tan, Sixiang
               and Lin, Jiangzhaung and Bu, Wenxiu",
  journal   = "PLoS One",
  publisher = "Public Library of Science (PLoS)",
  volume    =  17,
  number    =  1,
  pages     = "e0262689",
  month     =  jan,
  year      =  2022,
  copyright = "http://creativecommons.org/licenses/by/4.0/",
  language  = "en"
}

@article{WANG2023104976,
title = {DHUnet: Dual-branch hierarchical global–local fusion network for whole slide image segmentation},
journal = {Biomedical Signal Processing and Control},
volume = {85},
pages = {104976},
year = {2023},
issn = {1746-8094},
doi = {https://doi.org/10.1016/j.bspc.2023.104976},
url = {https://www.sciencedirect.com/science/article/pii/S1746809423004093},
author = {Lian Wang and Liangrui Pan and Hetian Wang and Mingting Liu and Zhichao Feng and Pengfei Rong and Zuo Chen and Shaoliang Peng}
}

@article{DBLP:journals/corr/abs-2201-09792,
  author       = {Asher Trockman and
                  J. Zico Kolter},
  title        = {Patches Are All You Need?},
  journal      = {CoRR},
  volume       = {abs/2201.09792},
  year         = {2022},
  url          = {https://arxiv.org/abs/2201.09792},
  eprinttype    = {arXiv},
  eprint       = {2201.09792},
  bibsource    = {dblp computer science bibliography, https://dblp.org}
}

@article{DBLP:journals/corr/abs-1708-02002,
  author       = {Tsung{-}Yi Lin and
                  Priya Goyal and
                  Ross B. Girshick and
                  Kaiming He and
                  Piotr Doll{\'{a}}r},
  title        = {Focal Loss for Dense Object Detection},
  journal      = {CoRR},
  volume       = {abs/1708.02002},
  year         = {2017},
  url          = {http://arxiv.org/abs/1708.02002},
  eprinttype    = {arXiv},
  eprint       = {1708.02002},
  bibsource    = {dblp computer science bibliography, https://dblp.org}
}

@misc{Mappingt95:online,
author = {Minyoung E. Kim},
  title = {Mapping the Cellular Landscape of the Brain: A Scalable Approach to Comprehensive Microscopy Data Analysis},
  url = "https://dspace.mit.edu/handle/1721.1/157135?show=full",
month = {},
year = {},
  note = "[Online; accessed 2025-08-28]"
}

@UNPUBLISHED{Swaney2019-vh,
  title       = "Scalable image processing techniques for quantitative analysis
                 of volumetric biological images from light-sheet microscopy",
  author      = "Swaney, Justin and Kamentsky, Lee and Evans, Nicholas B and
                 Xie, Katherine and Park, Young-Gyun and Drummond, Gabrielle
                 and Yun, Dae Hee and Chung, Kwanghun",
  journal     = "bioRxiv",
  institution = "bioRxiv",
  month       =  mar,
  year        =  2019
}

@software{neuroglancer2021,
  title        = {Neuroglancer: WebGL-based viewer for volumetric data},
  author       = {Maitin-Shepard, Jeremy and Baden, Alex and Silversmith, William and Perlman, Eric and Collman, Forrest and Blakely, Tim and Funke, Jan and others},
  year         = {2021},
  version      = {v2.23},
  publisher    = {Zenodo},
  doi          = {10.5281/zenodo.5573294},
  url          = {https://doi.org/10.5281/zenodo.5573294}
}

@software{zarr2025,
  title = {{Zarr}: Chunked, compressed, N-dimensional arrays for Python},
  author = {{Zarr Developers}},
  year = {2025},
  version = {3.1.2},
  url = {https://pypi.org/project/zarr/},
  note = {Released August 25, 2025},
}

@ARTICLE{Dean2015-ft,
  title     = "Deconvolution-free subcellular imaging with axially swept light
               sheet microscopy",
  author    = "Dean, Kevin M and Roudot, Philippe and Welf, Erik S and Danuser,
               Gaudenz and Fiolka, Reto",
  journal   = "Biophys. J.",
  publisher = "Elsevier BV",
  volume    =  108,
  number    =  12,
  pages     = "2807--2815",
  month     =  jun,
  year      =  2015,
  copyright = "http://creativecommons.org/licenses/by-nc-nd/4.0/",
  language  = "en"
}

@article{bentley1975multidimensional,
  title={Multidimensional binary search trees used for associative searching},
  author={Bentley, Jon Louis},
  journal={Communications of the ACM},
  volume={18},
  number={9},
  pages={509--517},
  year={1975},
  publisher={ACM}
}

@article{Hwang2018,
  title="Single-cell RNA sequencing technologies and bioinformatics pipelines",
author="Hwang, B. and Lee, J.H. and Bang, D",
journal = "Experimental \& Molecular Medicine",
volume = 50,
pages = "1--4",
year = 2018}

@article{merfish,
  title="Spatially resolved, highly multiplexed RNA profiling in single cells",
author = "Chen, Kok Hao and Boettiger, Alistair N. and Moffitt, Jeffrey R. and Wang, Siyuan and Zhuang,Xiaowei",
journal = "Science",
volume = 348,
year = 2015}

@article{mrnaseq,
title="mRNA-Seq whole-transcriptome analysis of a single cell",
author="Tang, F. and Barbacioru, C. and Wang, Y. and others",
journal="Nature Methods",
volume = 6,
pages="377--382",
year=2009}

@article{ku2020,
title="Elasticizing tissues for reversible shape transformation and accelerated molecular labeling",
author="Ku, Taeyun and Guan, Webster and Evans, Nicholas B. and Sohn, Chang Ho  and Albanese, Alexandre and Kim, Joon-Goon and Frosch, Matthew P.  and Chung,Kwanghun",
journal="Nature Methods",
volume = 17,
pages="609--613",
year=2020}

@article{park2024science,
title="Integrated platform for multiscale molecular imaging and phenotyping of the human brain",
author="Park, Juhyuk and Wang, Ji and Guan, Webster and others",
journal="Science",
volume=384,
year=2024}

@article{mri,
title="Overview of functional magnetic resonance imaging",
author="Glover, Gary H.",
journal="Neurosurgery Clinics of North America",
year=2012,
volume=22,
number=2,
pages="133--139"}

@article{zheng2018,
title="A Complete Electron Microscopy Volume of the Brain of Adult Drosophila melanogaster",
author="Zheng, Zhihao  and Lauritzen, J. Scott and Perlman, Eric and others",
journal="Cell",
volume=174,
number=3,
pages="730--743",
year=2018}

@article{Kasthuri2015,
title="Saturated Reconstruction of a Volume of Neocortex",
author="Kasthuri, Narayanan and Hayworth, Kenneth Jeffrey and Berger, Daniel Raimund and others",
journal="Cell",
volume=162,
number=3,
pages="648--661",
year=2015}

@article{boergens2017webknossos,
  title        = {WEBKNOSSOS: Efficient Online 3D Data Annotation for Connectomics},
  author       = {Boergens, K. M. and Berning, M. and Bocklisch, T. and Br{\"a}unlein, D. and Drawitsch, F. and Frohnhofen, J. and Herold, T. and Otto, P. and Rzepka, N. and Werkmeister, T. and Werner, D. and Wiese, G. and Wissler, H. and Helmstaedter, M.},
  journal      = {Nature Methods},
  volume       = {14},
  number       = {7},
  pages        = {691--694},
  year         = {2017},
  publisher    = {Nature Publishing Group},
  doi          = {10.1038/nmeth.4302}
}

@misc{oquab2024dinov2learningrobustvisual,
      title={DINOv2: Learning Robust Visual Features without Supervision}, 
      author={Maxime Oquab and Timothée Darcet and Théo Moutakanni and Huy Vo and Marc Szafraniec and Vasil Khalidov and Pierre Fernandez and Daniel Haziza and Francisco Massa and Alaaeldin El-Nouby and Mahmoud Assran and Nicolas Ballas and Wojciech Galuba and Russell Howes and Po-Yao Huang and Shang-Wen Li and Ishan Misra and Michael Rabbat and Vasu Sharma and Gabriel Synnaeve and Hu Xu and Hervé Jegou and Julien Mairal and Patrick Labatut and Armand Joulin and Piotr Bojanowski},
      year={2024},
      eprint={2304.07193},
      archivePrefix={arXiv},
      primaryClass={cs.CV},
      url={https://arxiv.org/abs/2304.07193}, 
}

@inproceedings{conf/nips/RenHGS15,
  author = {Ren, Shaoqing and He, Kaiming and Girshick, Ross B. and Sun, Jian},
  booktitle = {NIPS},
  editor = {Cortes, Corinna and Lawrence, Neil D. and Lee, Daniel D. and Sugiyama, Masashi and Garnett, Roman},
  ee = {http://papers.nips.cc/paper/5638-faster-r-cnn-towards-real-time-object-detection-with-region-proposal-networks},
  pages = {91-99},
  title = {Faster R-CNN: Towards Real-Time Object Detection with Region Proposal Networks.},
  url = {http://dblp.uni-trier.de/db/conf/nips/nips2015.html#RenHGS15},
  year = 2015
}

@article{DBLP:journals/corr/RedmonDGF15,
  author       = {Joseph Redmon and
                  Santosh Kumar Divvala and
                  Ross B. Girshick and
                  Ali Farhadi},
  title        = {You Only Look Once: Unified, Real-Time Object Detection},
  journal      = {CoRR},
  volume       = {abs/1506.02640},
  year         = {2015},
  url          = {http://arxiv.org/abs/1506.02640},
  eprinttype    = {arXiv},
  eprint       = {1506.02640},
  bibsource    = {dblp computer science bibliography, https://dblp.org}
}

@article{DBLP:journals/corr/abs-2005-12872,
  author       = {Nicolas Carion and
                  Francisco Massa and
                  Gabriel Synnaeve and
                  Nicolas Usunier and
                  Alexander Kirillov and
                  Sergey Zagoruyko},
  title        = {End-to-End Object Detection with Transformers},
  journal      = {CoRR},
  volume       = {abs/2005.12872},
  year         = {2020},
  url          = {https://arxiv.org/abs/2005.12872},
  eprinttype    = {arXiv},
  eprint       = {2005.12872},
  bibsource    = {dblp computer science bibliography, https://dblp.org}
}

@article{DBLP:journals/corr/abs-2111-06377,
  author       = {Kaiming He and
                  Xinlei Chen and
                  Saining Xie and
                  Yanghao Li and
                  Piotr Doll{\'{a}}r and
                  Ross B. Girshick},
  title        = {Masked Autoencoders Are Scalable Vision Learners},
  journal      = {CoRR},
  volume       = {abs/2111.06377},
  year         = {2021},
  url          = {https://arxiv.org/abs/2111.06377},
  eprinttype    = {arXiv},
  eprint       = {2111.06377},
  bibsource    = {dblp computer science bibliography, https://dblp.org}
}

@misc{wandb,
title = {Experiment Tracking with Weights and Biases},
year = {2020},
note = {Software available from wandb.com},
url={https://www.wandb.com/},
author = {Biewald, Lukas},
}

@article{Munch2009,
  author    = {Beat M{\"u}nch and Pavel Trtik and Federica Marone and Marco Stampanoni},
  title     = {Stripe and ring artifact removal with combined wavelet--Fourier filtering},
  journal   = {Optics Express},
  volume    = {17},
  number    = {10},
  pages     = {8567--8591},
  year      = {2009},
  publisher = {Optica Publishing Group},
  doi       = {10.1364/OE.17.008567},
  url       = {https://opg.optica.org/oe/fulltext.cfm?uri=oe-17-10-8567}
}

@article{Archit2025muSAM,
  author  = {Archit, Anwai and Freckmann, Luca and Nair, Sushmita and Khalid, Nabeel and Hilt, Philipp and Rajashekar, Vikas and Freitag, Marei and Teuber, Carolin and Spitzner, Melanie and Tapia Contreras, Constanza and Buckley, Genevieve and von Haaren, Sebastian and Gupta, Sagnik and Grade, Marian and Wirth, Matthias and Schneider, G{\"u}nter and Dengel, Andreas and Ahmed, Sheraz and Pape, Constantin},
  title   = {Segment Anything for Microscopy},
  journal = {Nature Methods},
  volume  = {22},
  number  = {3},
  pages   = {579--591},
  year    = {2025},
  doi     = {10.1038/s41592-024-02580-4},
  url     = {https://www.nature.com/articles/s41592-024-02580-4}
}

@misc{pachitariu2025cellposesam,
  title        = {Cellpose-SAM: Superhuman Generalization for Cellular Segmentation},
  author       = {Pachitariu, Marius and Rariden, Michael and Stringer, Carsen},
  year         = {2025},
  howpublished = {bioRxiv},
  doi          = {10.1101/2025.04.28.651001}
}

@article{Kirillov2018,
  author  = {Kirillov, Alexander and He, Kaiming and Girshick, Ross and Rother, Carsten and Doll{\'a}r, Piotr},
  title   = {Panoptic Segmentation},
  journal = {Proceedings of the IEEE/CVF Conference on Computer Vision and Pattern Recognition (CVPR)},
  year    = {2019},
  pages   = {9404--9413},
  doi     = {10.1109/CVPR.2019.00963},
  eprint  = {1801.00868},
  archivePrefix = {arXiv},
  primaryClass  = {cs.CV}
}
}
\clearpage
\setcounter{page}{1}
\maketitlesupplementary

\section{Supplementary}
\label{sec:supp1}

\subsection{Data Acquisition}
\label{sec:supp_data_acquisition}
%------------------------------------------------------------------------
\subsubsection{Sample Processing}
To generate raw datasets for each marker, we used \textit{Swiss Webster} mice (\textit{8 wks+}, \textit{Females}) for immunolabeling experiments. The parvalbumin dataset was generated using a \textit{B6 Pvalb-IRES-Cre::tdTomato} mouse (\textit{8 wks}, \textit{Female}). The following primary antibodies were used for immunolabeling: anti-NeuN CST \#36662 (\textit{6 $\mu$g}), anti-TH Biolegend \#818001 (\textit{6 $\mu$g}), anti-c-Fos Abcam \#ab214672 (\textit{3.5 $\mu$g, no special manipulation}), anti-IBA1 CST \#79394 (\textit{6 $\mu$g}), and anti-GFAP Invitrogen \#13-0300 (\textit{6 $\mu$g}). 

For tissue processing, all incubations were performed in 20 mL of the respective solution with light shaking unless otherwise specified. Mice were first transcardically perfused with 4\% Paraformaldehyde (PFA), then extracted brains were fixed in 4\% PFA for additional 24 hours at 4°C. For SHIELD preservation\cite{park2019protection}, samples were incubated in SHIELD-OFF solution for 3 days at 4 °C, then in SHIELD-ON solution for 1 day at 37 °C. Samples were then delipidated in Clear+ Delipidation Buffer for 7 d at 45 °C. Whole brain immunolabeling steps were performed using eFLASH technology\cite{yun2025uniform}, which integrates stochastic electrotransport\cite{kim2015stochastic}, implemented on a SmartBatch+ device (LifeCanvas Technologies). Indirect immunolabeling was performed using 1:2 molar ratio between primary and secondary antibodies. Primary immunolabeling was performed using the RADIANT Buffer System (LifeCanvas Technologies), followed by secondary labeling using the SmartBatch+ Secondary Buffer System (LifeCanvas Technologies), with the device presets Labeling 1 at 24 hours and Labeling 2 at 12 hours, respectively. 

\subsubsection{Imaging}
For optical clearing, samples were incubated in 50\% EasyIndex in denionized water, then in 100\% EasyIndex (RI = 1.52; LifeCanvas Technologies) both for 1 day at 37 °C. Refractive index matched samples were imaged using a SmartSPIM axially swept light-sheet microscope (LifeCanvas Technologies) at 3.6× (0.2 NA) magnification. Regarding potential destriping artifacts, we have carefully tuned the FFT–wavelet filter over hundreds of samples to prioritize signal preservation, even if minor residual striping remains, and observe improved image quality\cite{Swaney2019-vh}.

\subsection{Baseline Models}
\subsubsection{Cell location prediction}

The valid detection satisfies \Cref{eq2} with the following algorithm.

\begin{algorithm}[h!]
\footnotesize
\caption{Model with Location Prediction}
\SetAlgoLined
\label{alg:algo2}
\SetKwInOut{Input}{Input}
\SetKwInOut{Output}{Output}
\SetKwInOut{Parameters}{Parameters}
\SetKwComment{Comment}{/* }{ */}
\SetKw{Return}{return}
\SetKwFunction{Cast}{cast}
\SetKwFunction{MaxPoolThreeD}{MaxPool3D}
\SetKwFunction{ZeroPadThreeD}{ZeroPad3D}
\SetKwFunction{FindMaxima}{FindMaxima}
\SetKwFunction{Model}{Model}
\SetKwFunction{InputLayer}{InputLayer}
\Input{Detection Model $M$, threshold $\tau \in [0,1]$}
\Output{Location prediction model $M_{loc}$}
\Parameters{$d_{min} = 3$ (minimum distance between peaks)}

\BlankLine
    $I_{img} \gets\text{Volumetric Image Input from Model $M$}$
    
    $I_{idx} \gets \text{Batch index}$
    
    $H \gets \text{Model output; 3D probability heatmap}$
    
    $k \gets 2 \times d_{min} + 1$ 
    
    $H_{max} \gets \MaxPoolThreeD(H, \text{kernel}=(k, k, k), \text{stride}=(1, 1, 1), \text{padding}=\text{"valid"})$ 

    $H_{max} \gets \ZeroPadThreeD(H_{max}, \text{padding}=d_{min})$

    $L \gets \FindMaxima(H, H_{max}, I_{idx}, \tau)$
    
    $M_{loc} \gets \Model(I_{img}, I_{idx}, L)$ 

    \Return $M_{loc}$
\end{algorithm}

\subsubsection{Cross-dataset performance of baseline models}
Table \ref{tab:model_results} shows comprehensive evaluation of each baseline model across all six datasets. In general, models performed best on their corresponding cell type, confirming dataset-specific specialization. However, notable exceptions exist: the GFAP model performed poorly on its own dataset (F1=0.33), where the Iba1 model substantially outperformed it (F1=0.61), likely due to GFAP having the fewest annotations (2,530 total GT cells) while the Iba1 model benefits from a significantly larger training set (6,170 GT cells). Similarly, for some PV and cFos regions, non-matched models occasionally achieved comparable or higher F1 scores. Bold indicates evaluation on the matched dataset; underline indicates the highest-performing model for each region.

%%%%%%%%%%%%%%%%%%%%%%%%%%%%%%%%%%%%%%%%%%
% GIGANTIC TABLE
%%%%%%%%%%%%%%%%%%%%%%%%%%%%%%%%%%%%%%%%%%

% Requires:
% \usepackage{booktabs}
% \usepackage{multirow}
% \usepackage[table]{xcolor}
% \usepackage{adjustbox}

\begin{table*}[t]
\centering
\caption{Performance metrics (Accuracy and F1 Score) for six cell-type models across regions. (\textbf{Bold} indicates evaluation on matched data; \underline{underline} indicates the highest-performing model.)}
\label{tab:model_results}
\scriptsize
\setlength{\tabcolsep}{4pt}
\renewcommand{\arraystretch}{1.08}
\begin{adjustbox}{max width=\textwidth}
\begin{tabular}{llcccccccccccc}
\toprule
\multirow{2}{*}{\textbf{Cell Type}} & \multirow{2}{*}{\textbf{Region}}
& \multicolumn{2}{c}{\textbf{cFos Model}}
& \multicolumn{2}{c}{\textbf{NeuN Model}}
& \multicolumn{2}{c}{\textbf{TH Model}}
& \multicolumn{2}{c}{\textbf{PV Model}}
& \multicolumn{2}{c}{\textbf{GFAP Model}}
& \multicolumn{2}{c}{\textbf{Iba1 Model}} \\
\cmidrule(lr){3-4}
\cmidrule(lr){5-6}
\cmidrule(lr){7-8}
\cmidrule(lr){9-10}
\cmidrule(lr){11-12}
\cmidrule(lr){13-14}
& & Accuracy & F1\_Score
& Accuracy & F1\_Score
& Accuracy & F1\_Score
& Accuracy & F1\_Score
& Accuracy & F1\_Score
& Accuracy & F1\_Score \\
\midrule

\multirow{9}{*}{\textbf{cFos}}
& \cellcolor[gray]{0.90}\textbf{Total}
& \cellcolor[gray]{0.90}\underline{\textbf{0.64}} & \cellcolor[gray]{0.90}\underline{\textbf{0.78}} & \cellcolor[gray]{0.90}0.62 & \cellcolor[gray]{0.90}0.76 & \cellcolor[gray]{0.90}0.15 & \cellcolor[gray]{0.90}0.26 & \cellcolor[gray]{0.90}0.59 & \cellcolor[gray]{0.90}0.74 & \cellcolor[gray]{0.90}0.17 & \cellcolor[gray]{0.90}0.29 & \cellcolor[gray]{0.90}0.58 & \cellcolor[gray]{0.90}0.74 \\
& \cellcolor[gray]{0.90}\textbf{Train Set}
& \cellcolor[gray]{0.90}\textbf{0.63} & \cellcolor[gray]{0.90}\textbf{0.77} & \cellcolor[gray]{0.90}\underline{0.63} & \cellcolor[gray]{0.90}\underline{0.78} & \cellcolor[gray]{0.90}0.14 & \cellcolor[gray]{0.90}0.25 & \cellcolor[gray]{0.90}0.60 & \cellcolor[gray]{0.90}0.75 & \cellcolor[gray]{0.90}0.16 & \cellcolor[gray]{0.90}0.28 & \cellcolor[gray]{0.90}0.58 & \cellcolor[gray]{0.90}0.73 \\
& \hspace{1em}region\_1
& \underline{\textbf{0.66}} & \underline{\textbf{0.80}} & 0.62 & 0.76 & 0.13 & 0.23 & 0.60 & 0.75 & 0.18 & 0.31 & 0.56 & 0.72 \\
& \hspace{1em}region\_2
& \underline{\textbf{0.72}} & \underline{\textbf{0.84}} & 0.67 & 0.80 & 0.15 & 0.25 & 0.67 & 0.80 & 0.11 & 0.19 & 0.58 & 0.73 \\
& \hspace{1em}region\_3
& \textbf{0.53} & \textbf{0.69} & \underline{0.62} & \underline{0.77} & 0.16 & 0.27 & 0.53 & 0.70 & 0.19 & 0.32 & 0.60 & 0.75 \\
& \cellcolor[gray]{0.90}\textbf{Test Set}
& \cellcolor[gray]{0.90}\underline{\textbf{0.64}} & \cellcolor[gray]{0.90}\underline{\textbf{0.78}} & \cellcolor[gray]{0.90}0.60 & \cellcolor[gray]{0.90}0.75 & \cellcolor[gray]{0.90}0.16 & \cellcolor[gray]{0.90}0.27 & \cellcolor[gray]{0.90}0.59 & \cellcolor[gray]{0.90}0.74 & \cellcolor[gray]{0.90}0.17 & \cellcolor[gray]{0.90}0.30 & \cellcolor[gray]{0.90}0.59 & \cellcolor[gray]{0.90}0.74 \\
& \hspace{1em}region\_4
& \underline{\textbf{0.66}} & \underline{\textbf{0.80}} & 0.60 & 0.75 & 0.15 & 0.25 & 0.55 & 0.71 & 0.18 & 0.30 & 0.57 & 0.72 \\
& \hspace{1em}region\_5
& \underline{\textbf{0.74}} & \underline{\textbf{0.85}} & 0.62 & 0.77 & 0.15 & 0.26 & 0.73 & 0.84 & 0.12 & 0.22 & 0.62 & 0.76 \\
& \hspace{1em}region\_6
& \textbf{0.54} & \textbf{0.70} & 0.58 & 0.73 & 0.18 & 0.31 & 0.52 & 0.69 & 0.21 & 0.35 & \underline{0.59} & \underline{0.74} \\
\midrule
\multirow{9}{*}{\textbf{NeuN}}
& \cellcolor[gray]{0.90}\textbf{Total}
& \cellcolor[gray]{0.90}0.01 & \cellcolor[gray]{0.90}0.02 & \cellcolor[gray]{0.90}\underline{\textbf{0.67}} & \cellcolor[gray]{0.90}\underline{\textbf{0.81}} & \cellcolor[gray]{0.90}0.12 & \cellcolor[gray]{0.90}0.21 & \cellcolor[gray]{0.90}0.57 & \cellcolor[gray]{0.90}0.73 & \cellcolor[gray]{0.90}0.02 & \cellcolor[gray]{0.90}0.04 & \cellcolor[gray]{0.90}0.40 & \cellcolor[gray]{0.90}0.57 \\
& \cellcolor[gray]{0.90}\textbf{Train Set}
& \cellcolor[gray]{0.90}0.01 & \cellcolor[gray]{0.90}0.02 & \cellcolor[gray]{0.90}\underline{\textbf{0.65}} & \cellcolor[gray]{0.90}\underline{\textbf{0.79}} & \cellcolor[gray]{0.90}0.11 & \cellcolor[gray]{0.90}0.20 & \cellcolor[gray]{0.90}0.57 & \cellcolor[gray]{0.90}0.72 & \cellcolor[gray]{0.90}0.02 & \cellcolor[gray]{0.90}0.04 & \cellcolor[gray]{0.90}0.38 & \cellcolor[gray]{0.90}0.55 \\
& \hspace{1em}region\_1
& 0.00 & 0.00 & \underline{\textbf{0.63}} & \underline{\textbf{0.77}} & 0.13 & 0.23 & 0.55 & 0.71 & 0.02 & 0.03 & 0.41 & 0.58 \\
& \hspace{1em}region\_2
& 0.07 & 0.13 & \underline{\textbf{0.53}} & \underline{\textbf{0.69}} & 0.13 & 0.23 & 0.38 & 0.55 & 0.06 & 0.11 & 0.31 & 0.48 \\
& \hspace{1em}region\_3
& 0.01 & 0.02 & \underline{\textbf{0.70}} & \underline{\textbf{0.83}} & 0.09 & 0.17 & 0.63 & 0.77 & 0.02 & 0.04 & 0.37 & 0.54 \\
& \cellcolor[gray]{0.90}\textbf{Test Set}
& \cellcolor[gray]{0.90}0.01 & \cellcolor[gray]{0.90}0.02 & \cellcolor[gray]{0.90}\underline{\textbf{0.69}} & \cellcolor[gray]{0.90}\underline{\textbf{0.82}} & \cellcolor[gray]{0.90}0.12 & \cellcolor[gray]{0.90}0.22 & \cellcolor[gray]{0.90}0.58 & \cellcolor[gray]{0.90}0.73 & \cellcolor[gray]{0.90}0.02 & \cellcolor[gray]{0.90}0.04 & \cellcolor[gray]{0.90}0.41 & \cellcolor[gray]{0.90}0.59 \\
& \hspace{1em}region\_4
& 0.00 & 0.00 & \underline{\textbf{0.73}} & \underline{\textbf{0.84}} & 0.13 & 0.23 & 0.56 & 0.72 & 0.02 & 0.03 & 0.38 & 0.55 \\
& \hspace{1em}region\_5
& 0.09 & 0.16 & \underline{\textbf{0.51}} & \underline{\textbf{0.67}} & 0.13 & 0.23 & 0.40 & 0.57 & 0.06 & 0.11 & 0.32 & 0.48 \\
& \hspace{1em}region\_6
& 0.01 & 0.02 & \underline{\textbf{0.70}} & \underline{\textbf{0.83}} & 0.12 & 0.21 & 0.64 & 0.78 & 0.02 & 0.04 & 0.48 & 0.64 \\
\midrule
\multirow{9}{*}{\textbf{TH}}
& \cellcolor[gray]{0.90}\textbf{Total}
& \cellcolor[gray]{0.90}0.02 & \cellcolor[gray]{0.90}0.04 & \cellcolor[gray]{0.90}0.25 & \cellcolor[gray]{0.90}0.41 & \cellcolor[gray]{0.90}\underline{\textbf{0.40}} & \cellcolor[gray]{0.90}\underline{\textbf{0.57}} & \cellcolor[gray]{0.90}0.11 & \cellcolor[gray]{0.90}0.20 & \cellcolor[gray]{0.90}0.08 & \cellcolor[gray]{0.90}0.14 & \cellcolor[gray]{0.90}0.16 & \cellcolor[gray]{0.90}0.28 \\
& \cellcolor[gray]{0.90}\textbf{Train Set}
& \cellcolor[gray]{0.90}0.01 & \cellcolor[gray]{0.90}0.02 & \cellcolor[gray]{0.90}0.25 & \cellcolor[gray]{0.90}0.40 & \cellcolor[gray]{0.90}\underline{\textbf{0.42}} & \cellcolor[gray]{0.90}\underline{\textbf{0.59}} & \cellcolor[gray]{0.90}0.06 & \cellcolor[gray]{0.90}0.12 & \cellcolor[gray]{0.90}0.07 & \cellcolor[gray]{0.90}0.13 & \cellcolor[gray]{0.90}0.16 & \cellcolor[gray]{0.90}0.28 \\
& \hspace{1em}region\_1
& 0.01 & 0.01 & 0.22 & 0.36 & \underline{\textbf{0.40}} & \underline{\textbf{0.57}} & 0.02 & 0.03 & 0.09 & 0.16 & 0.13 & 0.23 \\
& \hspace{1em}region\_2
& 0.00 & 0.01 & 0.11 & 0.21 & \underline{\textbf{0.35}} & \underline{\textbf{0.51}} & 0.03 & 0.06 & 0.02 & 0.04 & 0.07 & 0.13 \\
& \hspace{1em}region\_3
& 0.04 & 0.07 & 0.54 & 0.70 & \underline{\textbf{0.66}} & \underline{\textbf{0.79}} & 0.21 & 0.34 & 0.11 & 0.19 & 0.29 & 0.45 \\
& \cellcolor[gray]{0.90}\textbf{Test Set}
& \cellcolor[gray]{0.90}0.03 & \cellcolor[gray]{0.90}0.06 & \cellcolor[gray]{0.90}0.26 & \cellcolor[gray]{0.90}0.41 & \cellcolor[gray]{0.90}\underline{\textbf{0.38}} & \cellcolor[gray]{0.90}\underline{\textbf{0.55}} & \cellcolor[gray]{0.90}0.16 & \cellcolor[gray]{0.90}0.28 & \cellcolor[gray]{0.90}0.09 & \cellcolor[gray]{0.90}0.16 & \cellcolor[gray]{0.90}0.16 & \cellcolor[gray]{0.90}0.27 \\
& \hspace{1em}region\_4
& 0.01 & 0.02 & 0.18 & 0.31 & \underline{\textbf{0.35}} & \underline{\textbf{0.51}} & 0.03 & 0.07 & 0.08 & 0.14 & 0.09 & 0.17 \\
& \hspace{1em}region\_5
& 0.01 & 0.03 & 0.08 & 0.14 & \underline{\textbf{0.29}} & \underline{\textbf{0.45}} & 0.04 & 0.07 & 0.01 & 0.03 & 0.03 & 0.05 \\
& \hspace{1em}region\_6
& 0.08 & 0.14 & 0.64 & 0.78 & \underline{\textbf{0.71}} & \underline{\textbf{0.83}} & 0.55 & 0.71 & 0.20 & 0.34 & 0.36 & 0.53 \\
\midrule
\multirow{9}{*}{\textbf{PV}}
& \cellcolor[gray]{0.90}\textbf{Total}
& \cellcolor[gray]{0.90}0.16 & \cellcolor[gray]{0.90}0.28 & \cellcolor[gray]{0.90}\underline{0.81} & \cellcolor[gray]{0.90}\underline{0.89} & \cellcolor[gray]{0.90}0.52 & \cellcolor[gray]{0.90}0.68 & \cellcolor[gray]{0.90}\textbf{0.46} & \cellcolor[gray]{0.90}\textbf{0.63} & \cellcolor[gray]{0.90}0.12 & \cellcolor[gray]{0.90}0.21 & \cellcolor[gray]{0.90}0.45 & \cellcolor[gray]{0.90}0.62 \\
& \cellcolor[gray]{0.90}\textbf{Train Set}
& \cellcolor[gray]{0.90}0.17 & \cellcolor[gray]{0.90}0.29 & \cellcolor[gray]{0.90}\underline{0.81} & \cellcolor[gray]{0.90}\underline{0.89} & \cellcolor[gray]{0.90}0.50 & \cellcolor[gray]{0.90}0.67 & \cellcolor[gray]{0.90}\textbf{0.42} & \cellcolor[gray]{0.90}\textbf{0.59} & \cellcolor[gray]{0.90}0.11 & \cellcolor[gray]{0.90}0.20 & \cellcolor[gray]{0.90}0.48 & \cellcolor[gray]{0.90}0.65 \\
& \hspace{1em}region\_1
& 0.18 & 0.30 & \underline{0.67} & \underline{0.81} & 0.66 & 0.79 & \textbf{0.50} & \textbf{0.66} & 0.12 & 0.22 & 0.33 & 0.50 \\
& \hspace{1em}region\_2
& 0.00 & 0.00 & \underline{0.78} & \underline{0.87} & 0.33 & 0.49 & \textbf{0.23} & \textbf{0.37} & 0.03 & 0.07 & 0.43 & 0.60 \\
& \hspace{1em}region\_3
& 0.40 & 0.58 & \underline{0.99} & \underline{1.00} & 0.96 & 0.98 & \textbf{0.96} & \textbf{0.98} & 0.31 & 0.47 & 0.86 & 0.93 \\
& \cellcolor[gray]{0.90}\textbf{Test Set}
& \cellcolor[gray]{0.90}0.16 & \cellcolor[gray]{0.90}0.28 & \cellcolor[gray]{0.90}\underline{0.81} & \cellcolor[gray]{0.90}\underline{0.89} & \cellcolor[gray]{0.90}0.53 & \cellcolor[gray]{0.90}0.70 & \cellcolor[gray]{0.90}\textbf{0.51} & \cellcolor[gray]{0.90}\textbf{0.67} & \cellcolor[gray]{0.90}0.13 & \cellcolor[gray]{0.90}0.23 & \cellcolor[gray]{0.90}0.42 & \cellcolor[gray]{0.90}0.60 \\
& \hspace{1em}region\_4
& 0.15 & 0.27 & \underline{0.65} & \underline{0.79} & 0.58 & 0.73 & \textbf{0.53} & \textbf{0.70} & 0.12 & 0.22 & 0.27 & 0.42 \\
& \hspace{1em}region\_5
& 0.00 & 0.01 & \underline{0.79} & \underline{0.88} & 0.39 & 0.56 & \textbf{0.35} & \textbf{0.52} & 0.04 & 0.08 & 0.37 & 0.54 \\
& \hspace{1em}region\_6
& 0.39 & 0.56 & \underline{0.99} & \underline{1.00} & 0.95 & 0.98 & \textbf{0.97} & \textbf{0.98} & 0.36 & 0.53 & 0.87 & 0.93 \\
\midrule
\multirow{9}{*}{\textbf{GFAP}}
& \cellcolor[gray]{0.90}\textbf{Total}
& \cellcolor[gray]{0.90}0.00 & \cellcolor[gray]{0.90}0.00 & \cellcolor[gray]{0.90}0.03 & \cellcolor[gray]{0.90}0.05 & \cellcolor[gray]{0.90}0.02 & \cellcolor[gray]{0.90}0.04 & \cellcolor[gray]{0.90}0.00 & \cellcolor[gray]{0.90}0.01 & \cellcolor[gray]{0.90}\textbf{0.20} & \cellcolor[gray]{0.90}\textbf{0.33} & \cellcolor[gray]{0.90}\underline{0.44} & \cellcolor[gray]{0.90}\underline{0.61} \\
& \cellcolor[gray]{0.90}\textbf{Train Set}
& \cellcolor[gray]{0.90}0.00 & \cellcolor[gray]{0.90}0.00 & \cellcolor[gray]{0.90}0.01 & \cellcolor[gray]{0.90}0.03 & \cellcolor[gray]{0.90}0.01 & \cellcolor[gray]{0.90}0.03 & \cellcolor[gray]{0.90}0.00 & \cellcolor[gray]{0.90}0.01 & \cellcolor[gray]{0.90}\textbf{0.19} & \cellcolor[gray]{0.90}\textbf{0.32} & \cellcolor[gray]{0.90}\underline{0.41} & \cellcolor[gray]{0.90}\underline{0.58} \\
& \hspace{1em}region\_1
& 0.00 & 0.00 & 0.00 & 0.00 & 0.00 & 0.01 & 0.00 & 0.01 & \textbf{0.20} & \textbf{0.34} & \underline{0.48} & \underline{0.65} \\
& \hspace{1em}region\_2
& 0.00 & 0.00 & 0.03 & 0.05 & 0.03 & 0.06 & 0.00 & 0.00 & \textbf{0.15} & \textbf{0.26} & \underline{0.31} & \underline{0.48} \\
& \hspace{1em}region\_3
& 0.00 & 0.00 & 0.13 & 0.23 & 0.12 & 0.21 & 0.00 & 0.00 & \textbf{0.19} & \textbf{0.31} & \underline{0.23} & \underline{0.37} \\
& \cellcolor[gray]{0.90}\textbf{Test Set}
& \cellcolor[gray]{0.90}0.00 & \cellcolor[gray]{0.90}0.00 & \cellcolor[gray]{0.90}0.05 & \cellcolor[gray]{0.90}0.09 & \cellcolor[gray]{0.90}0.03 & \cellcolor[gray]{0.90}0.06 & \cellcolor[gray]{0.90}0.00 & \cellcolor[gray]{0.90}0.00 & \cellcolor[gray]{0.90}\textbf{0.22} & \cellcolor[gray]{0.90}\textbf{0.35} & \cellcolor[gray]{0.90}\underline{0.49} & \cellcolor[gray]{0.90}\underline{0.66} \\
& \hspace{1em}region\_4
& 0.00 & 0.00 & 0.00 & 0.01 & 0.01 & 0.03 & 0.00 & 0.01 & \textbf{0.18} & \textbf{0.30} & \underline{0.48} & \underline{0.65} \\
& \hspace{1em}region\_5
& 0.00 & 0.00 & 0.12 & 0.21 & 0.05 & 0.10 & 0.00 & 0.01 & \textbf{0.23} & \textbf{0.37} & \underline{0.51} & \underline{0.67} \\
& \hspace{1em}region\_6
& 0.00 & 0.01 & 0.08 & 0.15 & 0.05 & 0.10 & 0.00 & 0.00 & \textbf{0.28} & \textbf{0.44} & \underline{0.50} & \underline{0.67} \\
\midrule
\multirow{9}{*}{\textbf{Iba1}}
& \cellcolor[gray]{0.90}\textbf{Total}
& \cellcolor[gray]{0.90}0.01 & \cellcolor[gray]{0.90}0.03 & \cellcolor[gray]{0.90}0.27 & \cellcolor[gray]{0.90}0.43 & \cellcolor[gray]{0.90}0.39 & \cellcolor[gray]{0.90}0.56 & \cellcolor[gray]{0.90}0.03 & \cellcolor[gray]{0.90}0.06 & \cellcolor[gray]{0.90}0.28 & \cellcolor[gray]{0.90}0.43 & \cellcolor[gray]{0.90}\underline{\textbf{0.69}} & \cellcolor[gray]{0.90}\underline{\textbf{0.81}} \\
& \cellcolor[gray]{0.90}\textbf{Train Set}
& \cellcolor[gray]{0.90}0.01 & \cellcolor[gray]{0.90}0.01 & \cellcolor[gray]{0.90}0.24 & \cellcolor[gray]{0.90}0.39 & \cellcolor[gray]{0.90}0.35 & \cellcolor[gray]{0.90}0.52 & \cellcolor[gray]{0.90}0.02 & \cellcolor[gray]{0.90}0.05 & \cellcolor[gray]{0.90}0.27 & \cellcolor[gray]{0.90}0.42 & \cellcolor[gray]{0.90}\underline{\textbf{0.64}} & \cellcolor[gray]{0.90}\underline{\textbf{0.78}} \\
& \hspace{1em}region\_1
& 0.02 & 0.03 & 0.41 & 0.59 & 0.47 & 0.64 & 0.05 & 0.09 & 0.30 & 0.46 & \underline{\textbf{0.72}} & \underline{\textbf{0.84}} \\
& \hspace{1em}region\_2
& 0.00 & 0.01 & 0.18 & 0.30 & 0.37 & 0.54 & 0.02 & 0.03 & 0.21 & 0.35 & \underline{\textbf{0.56}} & \underline{\textbf{0.72}} \\
& \hspace{1em}region\_3
& 0.00 & 0.00 & 0.12 & 0.22 & 0.21 & 0.34 & 0.01 & 0.02 & 0.29 & 0.45 & \underline{\textbf{0.65}} & \underline{\textbf{0.79}} \\
& \cellcolor[gray]{0.90}\textbf{Test Set}
& \cellcolor[gray]{0.90}0.02 & \cellcolor[gray]{0.90}0.04 & \cellcolor[gray]{0.90}0.30 & \cellcolor[gray]{0.90}0.46 & \cellcolor[gray]{0.90}0.43 & \cellcolor[gray]{0.90}0.60 & \cellcolor[gray]{0.90}0.03 & \cellcolor[gray]{0.90}0.07 & \cellcolor[gray]{0.90}0.29 & \cellcolor[gray]{0.90}0.44 & \cellcolor[gray]{0.90}\underline{\textbf{0.73}} & \cellcolor[gray]{0.90}\underline{\textbf{0.85}} \\
& \hspace{1em}region\_4
& 0.01 & 0.03 & 0.43 & 0.60 & 0.46 & 0.63 & 0.06 & 0.11 & 0.32 & 0.48 & \underline{\textbf{0.74}} & \underline{\textbf{0.85}} \\
& \hspace{1em}region\_5
& 0.01 & 0.03 & 0.32 & 0.48 & 0.46 & 0.63 & 0.03 & 0.05 & 0.23 & 0.38 & \underline{\textbf{0.65}} & \underline{\textbf{0.78}} \\
& \hspace{1em}region\_6
& 0.03 & 0.06 & 0.18 & 0.31 & 0.37 & 0.54 & 0.02 & 0.03 & 0.30 & 0.46 & \underline{\textbf{0.83}} & \underline{\textbf{0.91}} \\
\bottomrule
\end{tabular}
\end{adjustbox}
\end{table*}

\subsubsection{Preliminary evaluations of segmentation performance}
\label{sec:prelim_eval_seg_perf}

\begin{figure}[t]
  \centering
  \includegraphics[width=0.24\linewidth]{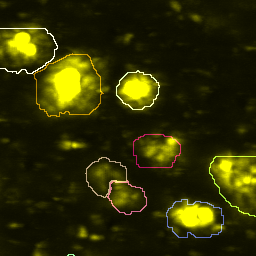}
  \includegraphics[width=0.24\linewidth]{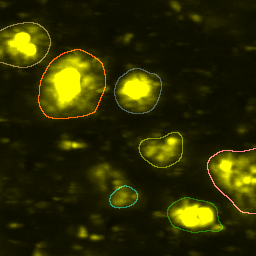}
  \includegraphics[width=0.24\linewidth]{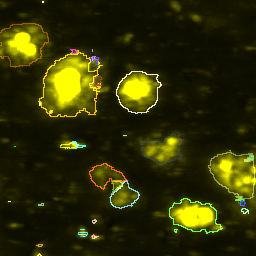}
  \includegraphics[width=0.24\linewidth]{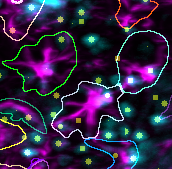}
  \setlength{\belowcaptionskip}{-10pt}
%  \caption{Example $\beta$-amyloid (yellow) segmentation results for (a, top left) ground truth, (b, top right) $\mu$SAM, and (c, bottom left) CellPose-SAM. (d, bottom right) $\mu$SAM segmentation overlaid with GFAP (magenta), nuclear stain (cyan), and detected nuclei cell centers (yellow).}
  \caption{Example $\beta$-amyloid (yellow) segmentation results for (a) ground truth, (b) $\mu$SAM, and (c) CellPose-SAM. (d) $\mu$SAM segmentation overlaid with GFAP (magenta), nuclear stain (cyan), and detected nuclei cell centers (yellow).}
  \label{fig:onecol}
\end{figure}

The downstream analyses enabled by the CANVAS dataset include regional cell density estimation, multi-channel co-expression analysis, and spatial proximity analysis between cell types. For these tasks, cell centroid annotations are generally sufficient and are substantially faster and more reliable to generate (particularly in large 3D datasets) than full instance segmentations. This is especially true for densely packed, irregularly shaped cells such as GFAP\textsuperscript{+} astrocytes, where overlapping processes make instance boundaries ambiguous and segmentation error-prone (Fig.~\ref{fig:onecol}d).

However, some biomarkers cannot be adequately represented by centroids alone. For example, $\beta$-amyloid plaques lack a meaningful “center” and require segmentation masks to capture their spatial extent and morphology. Segmentation also enables more detailed morphological characterization of cellular structures. To support this type of analysis, we are developing a new $\beta$-amyloid segmentation dataset based on ThioflavinS-stained adult AD mouse brains (\textit{5xFAD}) that will serve as a segmentation benchmark in future releases of CANVAS.

As a preliminary evaluation, we compared two recent SAM-based segmentation models (Cellpose-SAM \cite{pachitariu2025cellposesam} and $\mu$SAM \cite{Archit2025muSAM}) on $\beta$-amyloid plaques within a subsection of this dataset. Performance was evaluated using panoptic quality (PQ) \cite{Kirillov2018}, which jointly measures recognition quality (object-level F1) and segmentation quality (mean IoU over matched instances). A $200 \times 1000 \times 1000$ $(z,y,x)$ region was cropped for evaluation, and image intensities were clipped and normalized to the 99.5th percentile.

Qualitatively, $\mu$SAM was able to capture large $\beta$-amyloid plaques, whereas Cellpose-SAM frequently over-segmented plaques into smaller fragments (Fig.~\ref{fig:onecol}). This resulted in lower PQ for Cellpose-SAM (0.05) compared to $\mu$SAM (0.19). Although these models were not fine-tuned on our domain-specific data, the results suggest that SAM-based segmentation models already show promising performance on LSFM datasets. In future work, we plan to fine-tune models such as $\mu$SAM on the CANVAS dataset to further improve segmentation performance and establish stronger benchmarks for 3D biomarker segmentation.

\subsection{Representation learning with CANVAS}
\subsubsection{Model architecture}

For training the autoencoder part of the 3D-MAE model, we used a standard Vision Transformer architecture adapted for 3D volumetric data with the configuration shown in \Cref{table:3d-mae-architecture}, and hyperparameter settings shown in \Cref{table:3d-mae-hyperparams}.

\subsubsection{3D-MAE model training results}

\Cref{table:training_result_best_models} shows the 3-D MAE training results. Overall, lower mask ratios (0.15–0.55) outperformed the commonly used ratio of 0.75 reported in the original paper. Additionally, both Structural Similarity Index Measure (SSIM) and Peak Signal-to-Noise Ratio (PSNR) values for the best models are lower than those typically observed when training on natural images such as ImageNet. This is expected, as 3-D brain microscopy data contain sparse biological structures and SSIM is highly sensitive to structural patterns; with most patches being empty or nearly empty, SSIM provides limited meaningful signal. Nevertheless, the goal of this training is to learn useful representations for downstream tasks rather than to maximize reconstruction metrics.

Among the marker-specific trainings, NeuN achieved the lowest loss, consistent with its compact and relatively uniform cellular morphology. In contrast, GFAP reported the highest loss, reflecting the complexity and elongated processes of astrocytes, which make reconstruction more challenging. The all-markers model performed within 15\% of the best single-marker models despite capturing roughly six times greater morphological diversity.

\begin{table}[H]
\footnotesize
\centering
\caption{3D-MAE Model Configuration}
\begin{tabular}{lcc}
\toprule
Component & Configuration & Parameters \\
\midrule
\textbf{Encoder} & & 10.75M \\
\quad Hidden dimension & 384 & \\
\quad Number of layers & 6 & \\
\quad Attention heads & 6 & \\
\quad MLP ratio & 4 & \\
\midrule
\textbf{Decoder} & & 1.90M \\
\quad Hidden dimension & 192 & \\
\quad Number of layers & 4 & \\
\quad Attention heads & 3 & \\
\midrule
\textbf{Total} & & \textbf{12.65M} \\
\bottomrule
\end{tabular}
\label{table:3d-mae-architecture}
\end{table}

\begin{table}[H]
\footnotesize
\centering
\caption{Complete Hyperparameter Configuration}
\label{tab:hyperparams}
\begin{tabular}{ll}
\toprule
\textbf{Parameter} & \textbf{Value} \\
\midrule
\multicolumn{2}{l}{\textit{Architecture}} \\
Crop sizes & 16×32×32, 24×48×48, 32×64×64 \\
Patch sizes & 4×8×8, 6×12×12, 8×16×16 \\
Mask ratios & 0.15, 0.35, 0.55, 0.75 \\
\midrule
\multicolumn{2}{l}{\textit{Optimization}} \\
Optimizer & AdamW \\
Learning rate ($\eta$) & $1.5e^{-4}$ \\
Weight decay & 0.05 \\
Batch size & 64 \\
Epochs & 700 \\
LR schedule & Cosine annealing \\
Min LR & $0.01 \times \eta$ \\
Gradient clipping & $\|g\|_2 \leq 1.0$ \\
\midrule
\multicolumn{2}{l}{\textit{Data Augmentation}} \\
Intensity rescaling & [0, 1] normalization \\
Random flip & 0.5 \\
Random rotation & $\pm 10^\circ$ \\
Gaussian noise & $\sigma = 0.01$ \\
\midrule
\multicolumn{2}{l}{\textit{Content-Aware Weighting}} \\
Background weight & $w_{\text{bg}} = 1.0$ \\
Cell weight & $w_{\text{cell}} \approx 10.0$ \\
\bottomrule
\end{tabular}
\label{table:3d-mae-hyperparams}
\end{table}

\begin{table}[h]
\centering
\caption{3-MAE training results. Best configuration (16$\times$32$\times$32 / 4$\times$8$\times$8).}
\label{tab:mae_results}
\scriptsize
\begin{tabular}{lccccc}
\toprule
\textbf{Marker} & \textbf{Config} & \textbf{Mask} & \textbf{Loss} $\downarrow$ & \textbf{PSNR} $\uparrow$ & \textbf{SSIM} $\uparrow$ \\
\midrule
NeuN       & 16$\times$32$\times$32 / 4$\times$8$\times$8 & 0.15 & 0.0061 & 12.00 & 0.256 \\
cFos       & 16$\times$32$\times$32 / 4$\times$8$\times$8 & 0.55 & 0.0082 & 14.55 & 0.291 \\
TH         & 16$\times$32$\times$32 / 4$\times$8$\times$8 & 0.55 & 0.0143 & 13.38 & 0.295 \\
PV         & 16$\times$32$\times$32 / 4$\times$8$\times$8 & 0.15 & 0.0087 & 12.69 & 0.199 \\
IBA1       & 16$\times$32$\times$32 / 4$\times$8$\times$8 & 0.15 & 0.0103 & 15.61 & 0.194 \\
GFAP       & 16$\times$32$\times$32 / 4$\times$8$\times$8 & 0.55 & 0.0194 & 13.74 & 0.360 \\
\midrule
\textbf{all\_markers} & 16$\times$32$\times$32 / 4$\times$8$\times$8 & 0.15 & 0.0070 & 14.68 & 0.172 \\
\bottomrule
\end{tabular}
\label{table:training_result_best_models}
\end{table}

\subsubsection{Content-Aware Weighting Ablation}

To validate our content-aware reconstruction weighting ($w_\text{cell} = 10 \times w_\text{bg}$), we compared it with uniform weighting. Under uniform weighting, the model over-optimizes for trivial background reconstruction due to the high sparsity of our volumetric microscopy data. In contrast, content-aware weighting based on patch variance concentrates learning on cellular regions while preserving spatial context, yielding 5.1\% and 10.8\% improvements in PSNR and SSIM, respectively, for the NeuN data set (\Cref{table:ablation-ca-weighting}).

\begin{table}[H]
\footnotesize
\centering
\caption{Content-Aware Weighting Ablation (NeuN, 16×32×32 / 4×8×8, m=0.15)}
\begin{tabular}{lccc}
\toprule
\textbf{Weighting Strategy} & \textbf{Final Loss} & \textbf{PSNR (dB)} & \textbf{SSIM} \\
\midrule
Uniform ($w_i = 1.0$ for all) & 0.0068 & 11.42 & 0.231 \\
Content-aware ($w_i \in [1, 10]$) & \textbf{0.0061} & \textbf{12.00} & \textbf{0.256} \\
\midrule
Relative improvement & +11.5\% & +5.1\% & +10.8\% \\
\bottomrule
\end{tabular}
\label{table:ablation-ca-weighting}
\end{table}

%% =============================================================================
\subsubsection{Qualitative Results and Visualizations}
%% =============================================================================

\begin{figure}[H]
\centering
\includegraphics[width=0.98\textwidth]{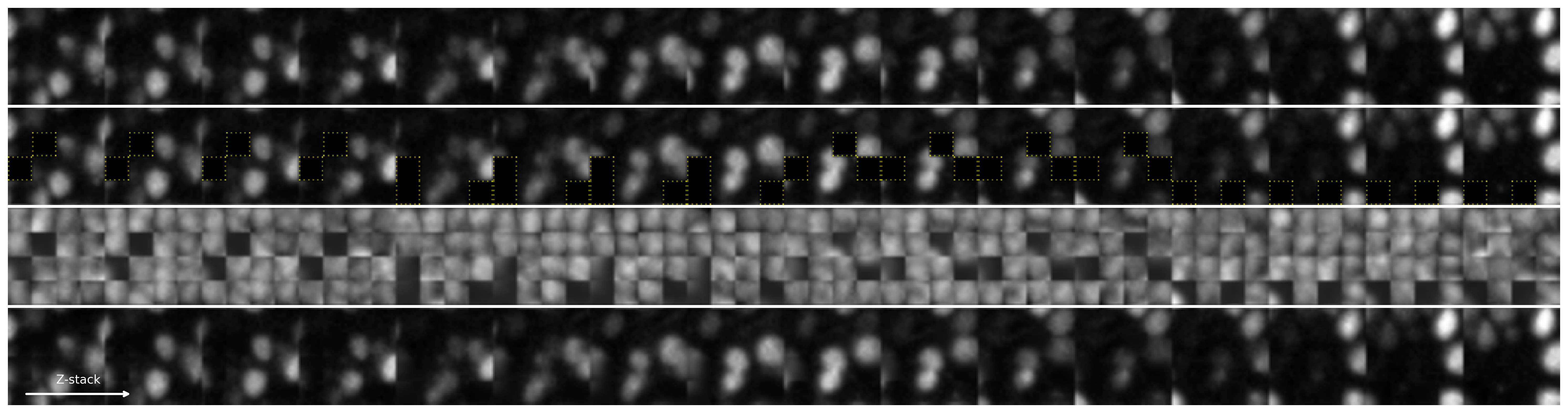}\\[0.5em]
\includegraphics[width=0.98\textwidth]{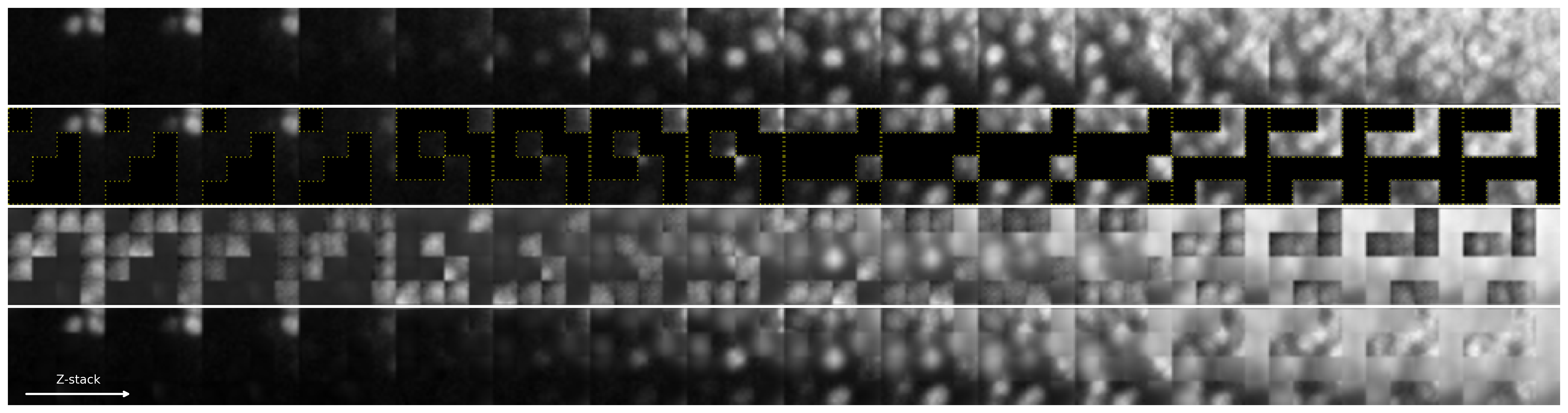}\\[0.5em]
\includegraphics[width=0.98\textwidth]{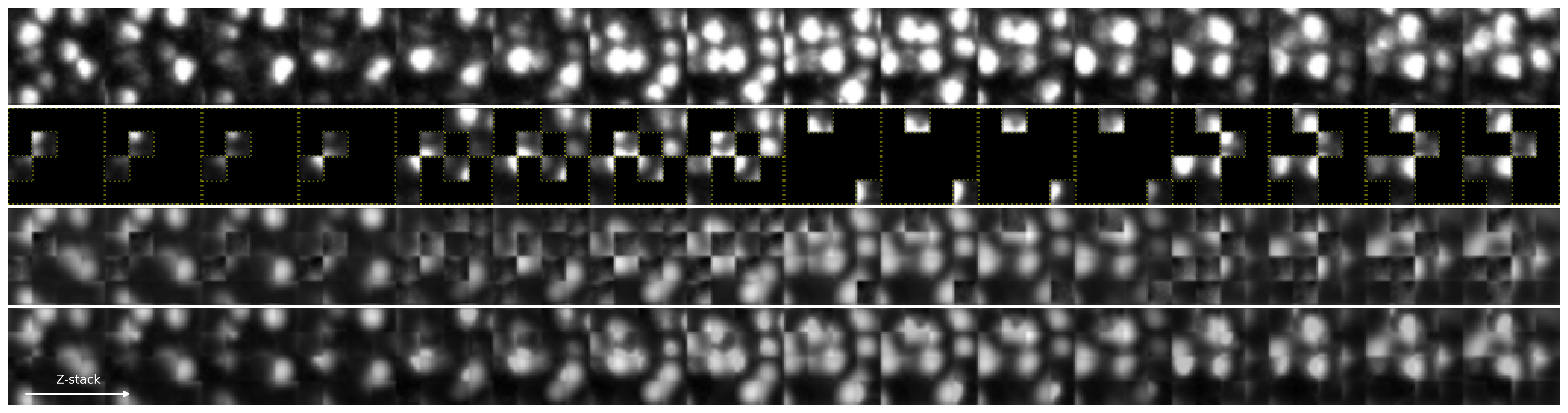}
\caption{\textbf{NeuN reconstruction with varying mask ratios.} Compact neuronal nuclei reconstructed with 16×32×32 crop, 4×8×8 patch. Top to bottom: mask ratio 0.15, 0.55, 0.75. Lower masking preserves more morphological detail.}
\label{fig:mae_recon_neun_masks}
\end{figure}

\begin{figure}[H]
\scriptsize
\centering
\includegraphics[width=0.98\textwidth]{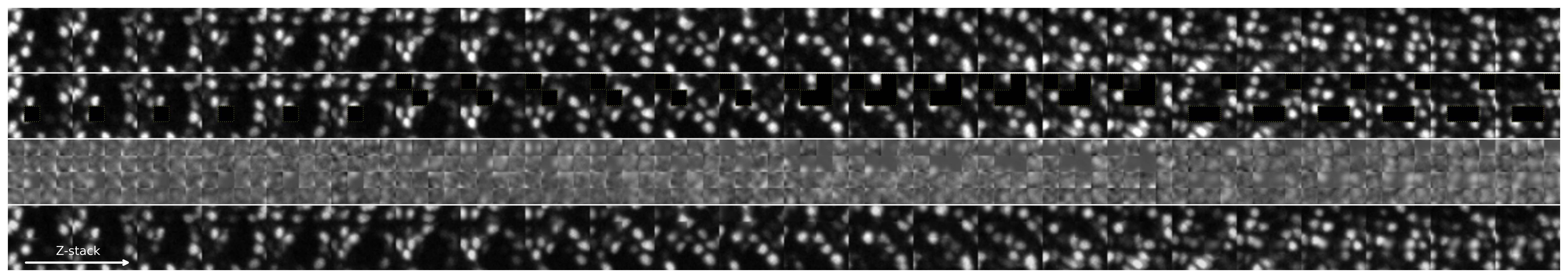}\\[0.5em]
\includegraphics[width=0.98\textwidth]{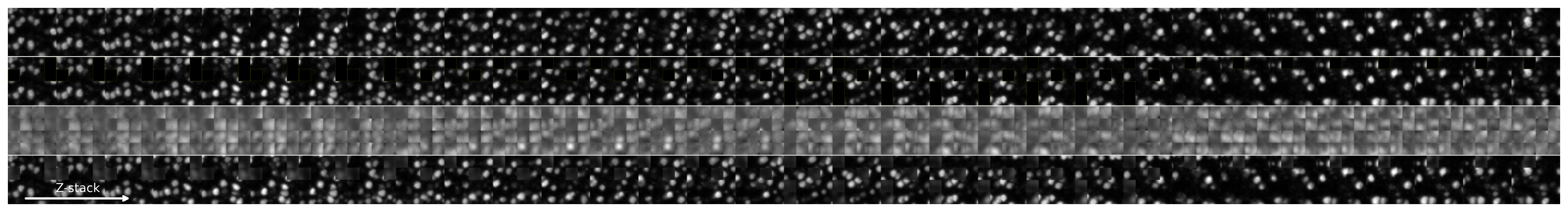}
\caption{\textbf{NeuN reconstruction with varying crop sizes.} Mask ratio 0.15. Top: 24×48×48 crop (6×12×12 patch). Bottom: 32×64×64 crop (8×16×16 patch). Larger crops capture more spatial context.}
\label{fig:mae_recon_neun_crops}
\end{figure}

\begin{figure}[H]
\centering
\includegraphics[width=0.98\textwidth]{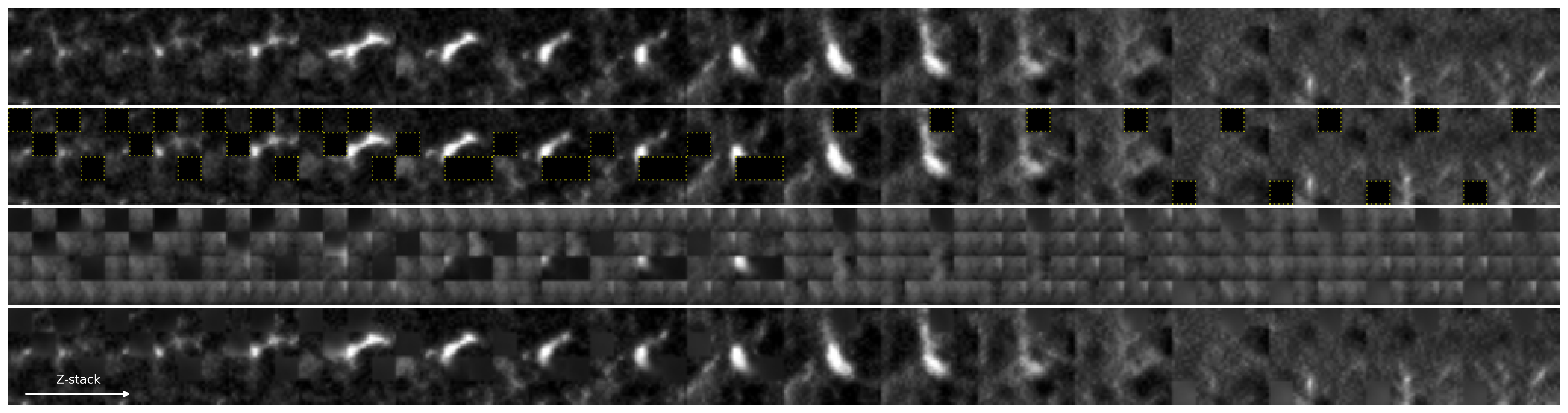}\\[0.5em]
\includegraphics[width=0.98\textwidth]{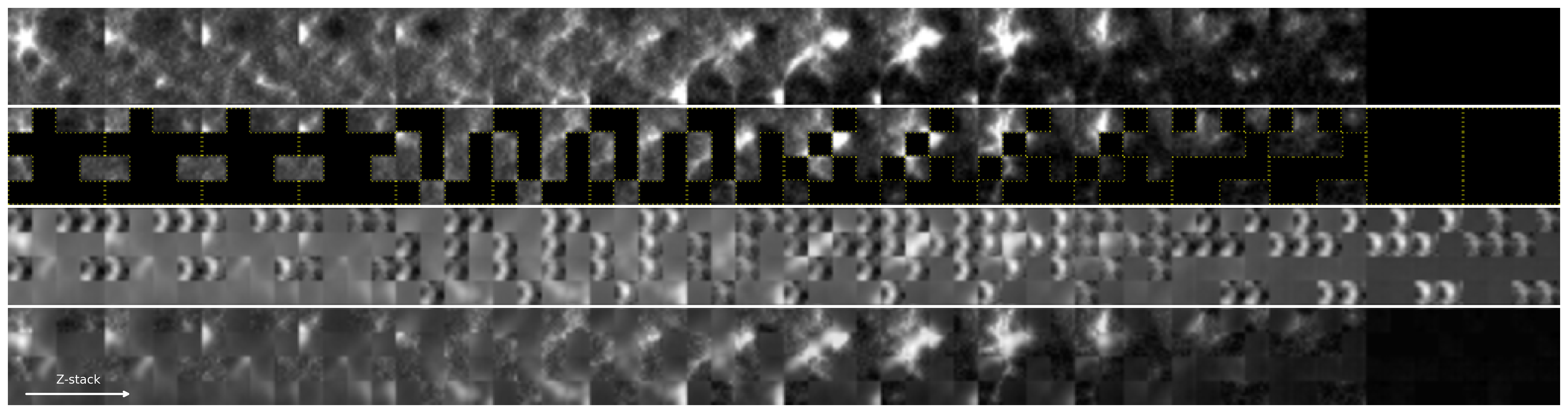}\\[0.5em]
\includegraphics[width=0.98\textwidth]{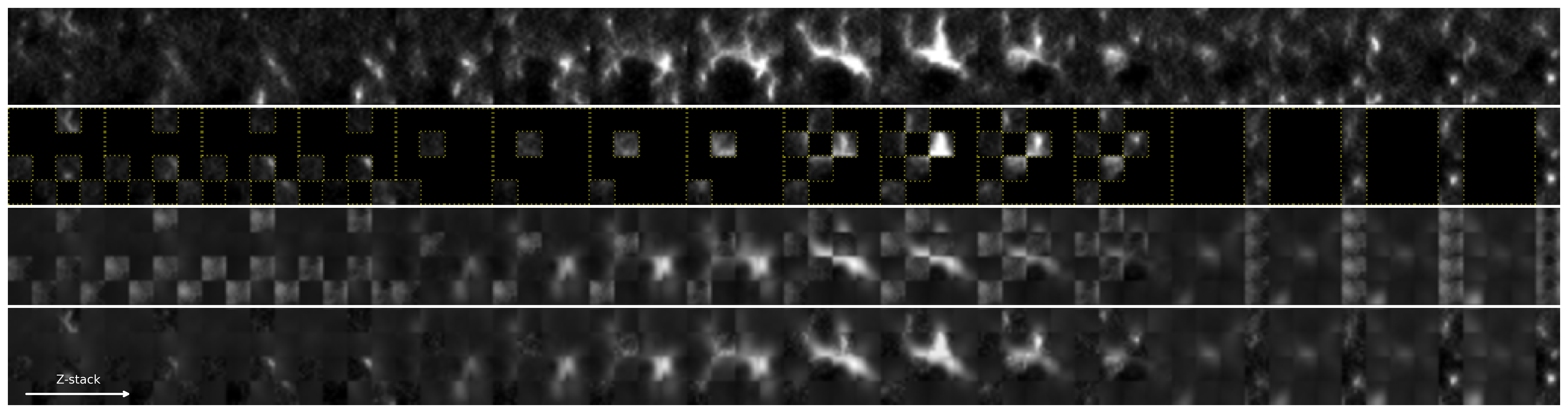}
\caption{\textbf{IBA1 reconstruction with varying mask ratios.} Ramified microglial morphology with fine processes. 16×32×32 crop, 4×8×8 patch.}
\label{fig:mae_recon_iba1}
\end{figure}

\begin{figure}[H]
\centering
\includegraphics[width=0.98\textwidth]{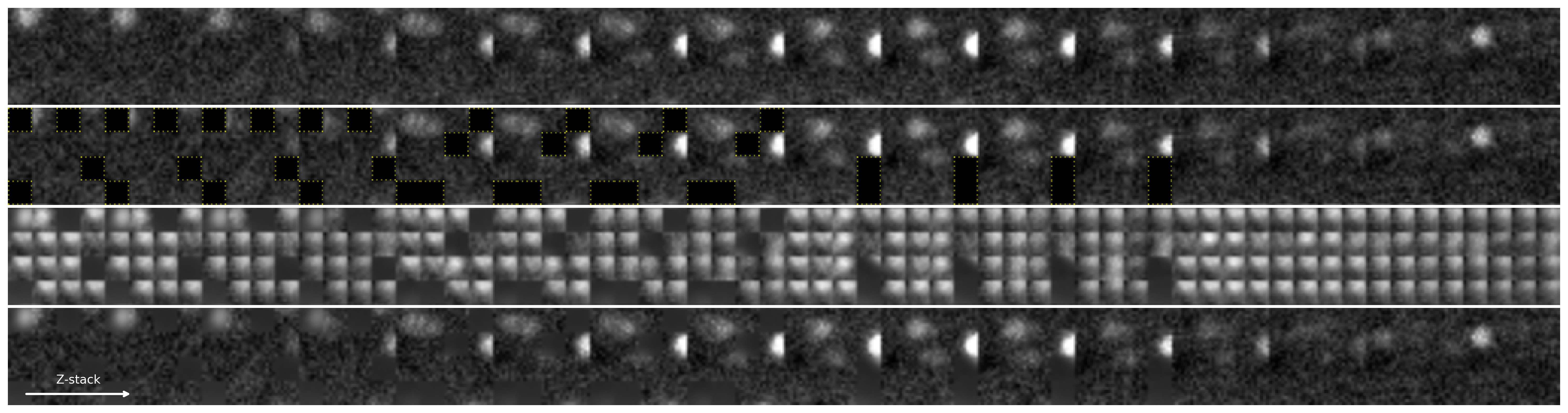}\\[0.5em]
\includegraphics[width=0.98\textwidth]{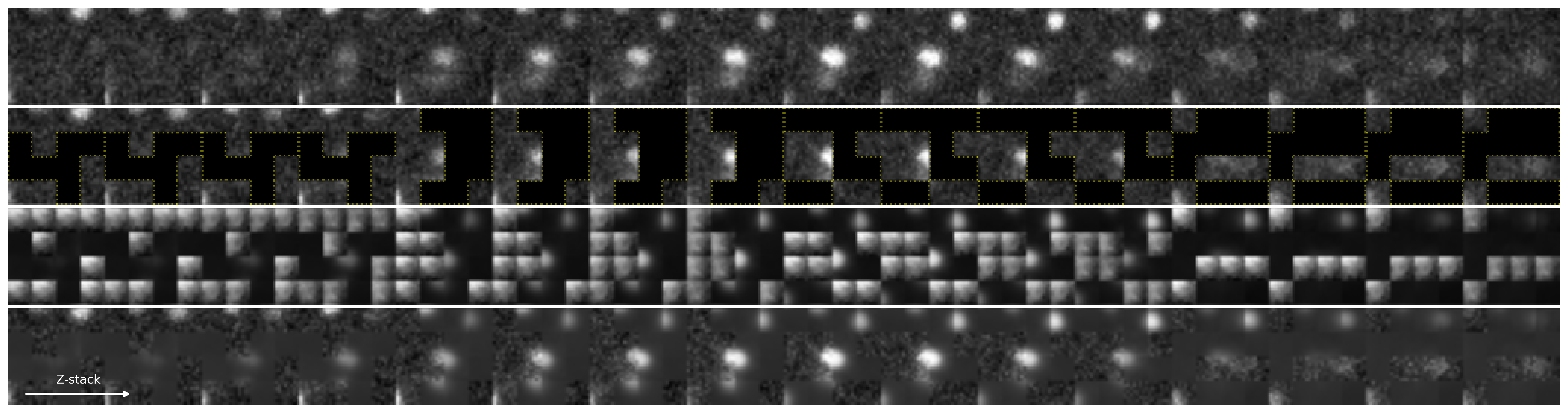}\\[0.5em]
\includegraphics[width=0.98\textwidth]{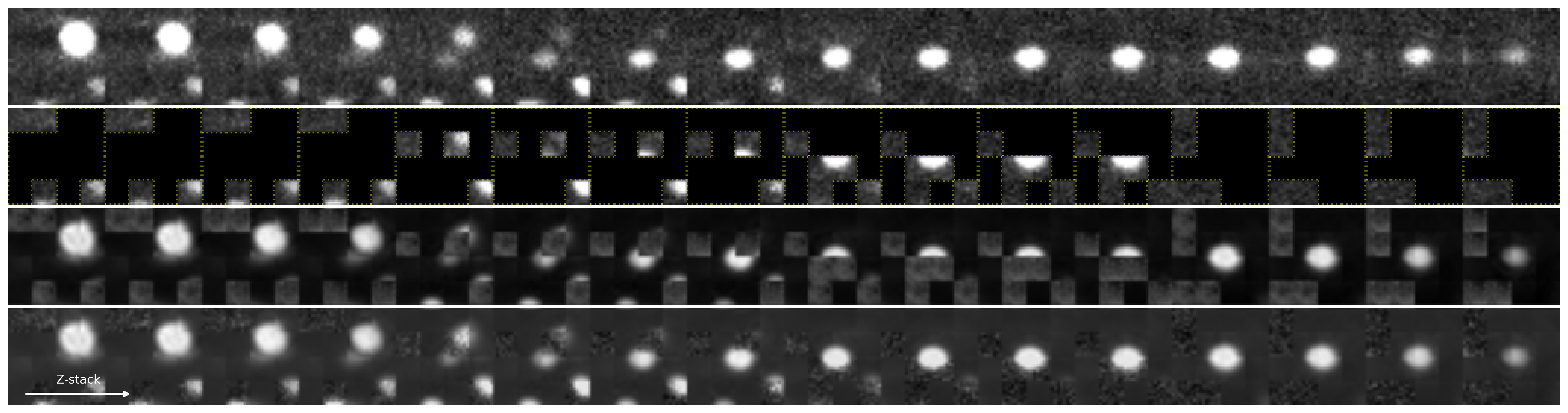}
\caption{\textbf{cFos reconstruction with varying mask ratios.} Sparse activity-dependent signal. 16×32×32 crop, 4×8×8 patch.}
\label{fig:mae_recon_cfos}
\end{figure}

\begin{figure}[H]
\centering
\includegraphics[width=0.98\textwidth]{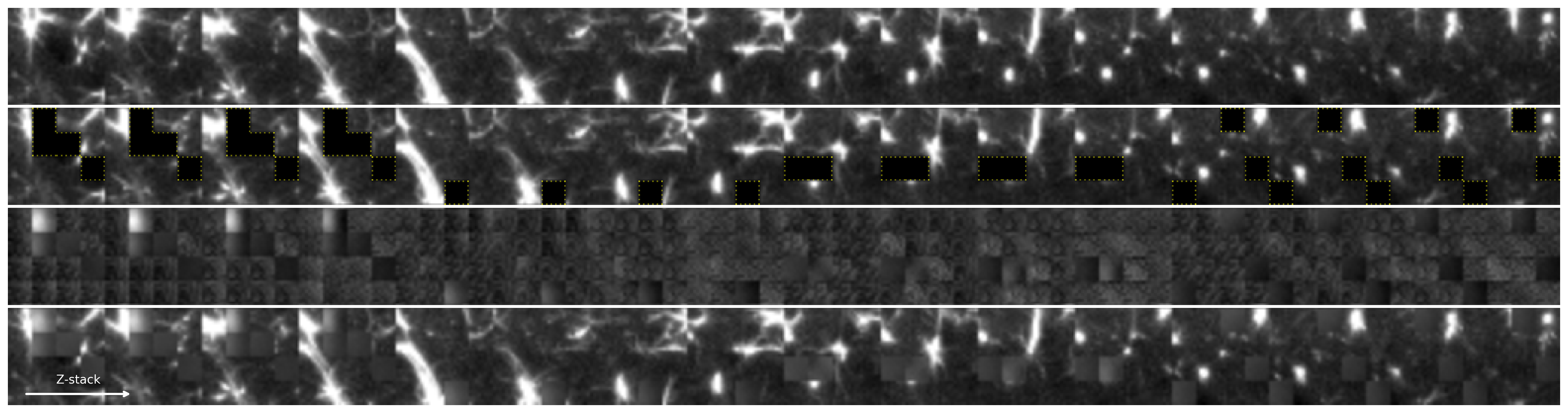}\\[0.5em]
\includegraphics[width=0.98\textwidth]{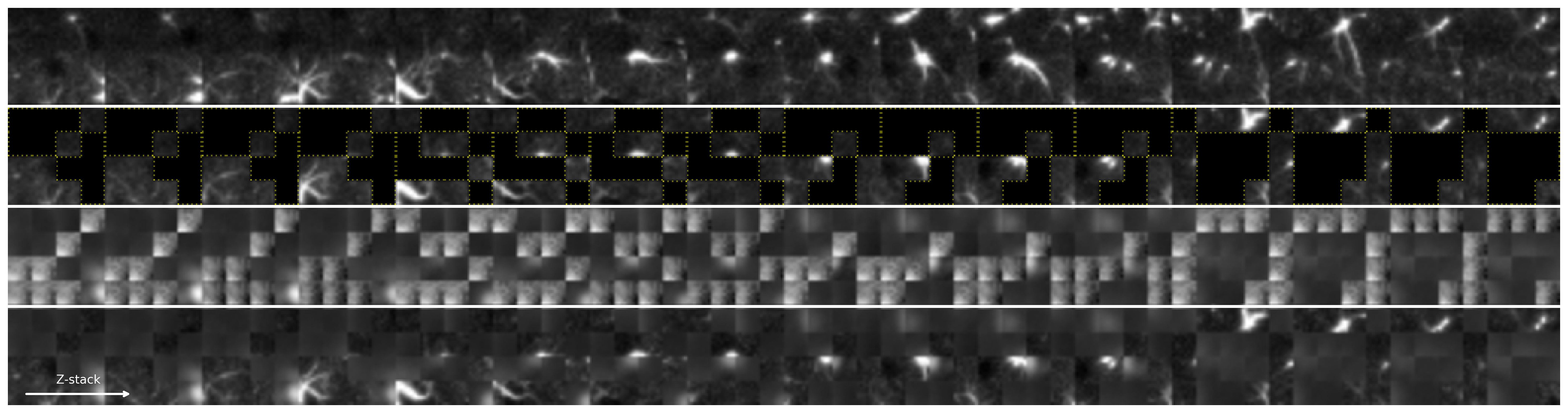}\\[0.5em]
\includegraphics[width=0.98\textwidth]{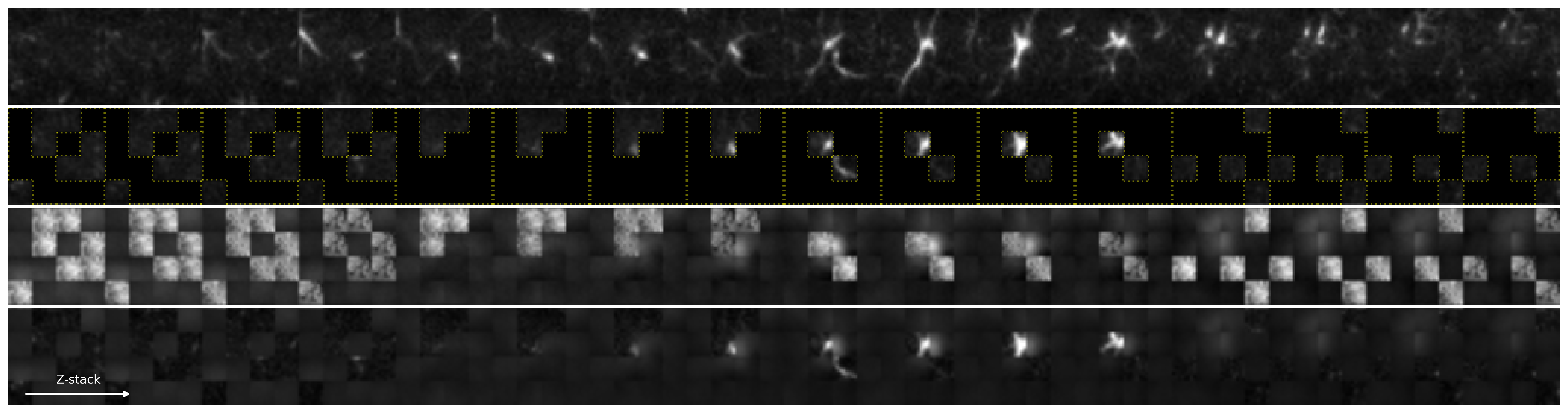}
\caption{\textbf{GFAP reconstruction with varying mask ratios.} Complex astrocyte morphology with elongated processes. 16×32×32 crop, 4×8×8 patch.}
\label{fig:mae_recon_gfap}
\end{figure}

\begin{figure}[H]
\centering
\includegraphics[width=0.98\textwidth]{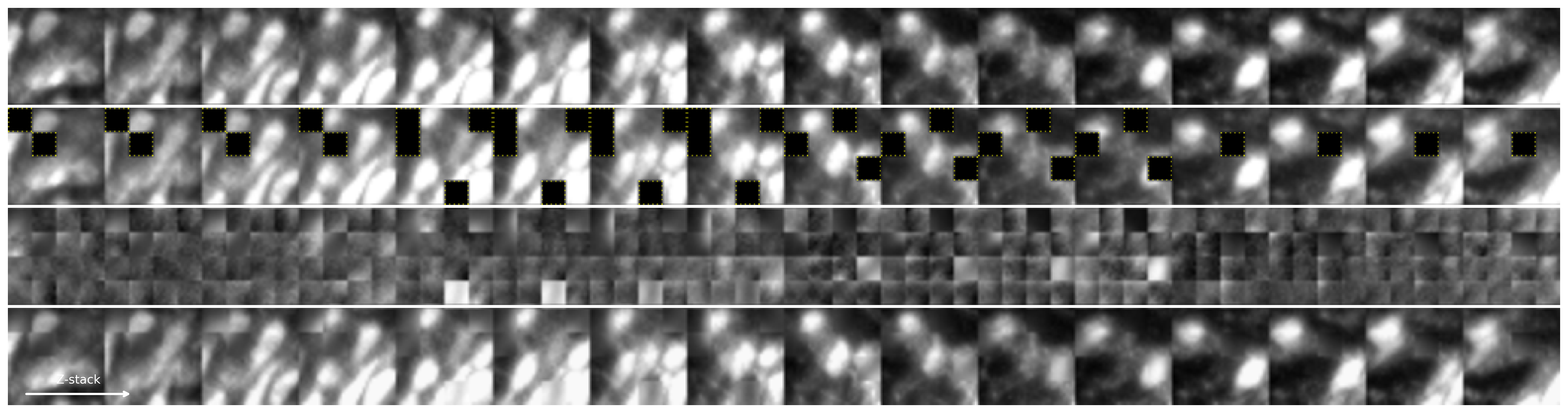}\\[0.5em]
\includegraphics[width=0.98\textwidth]{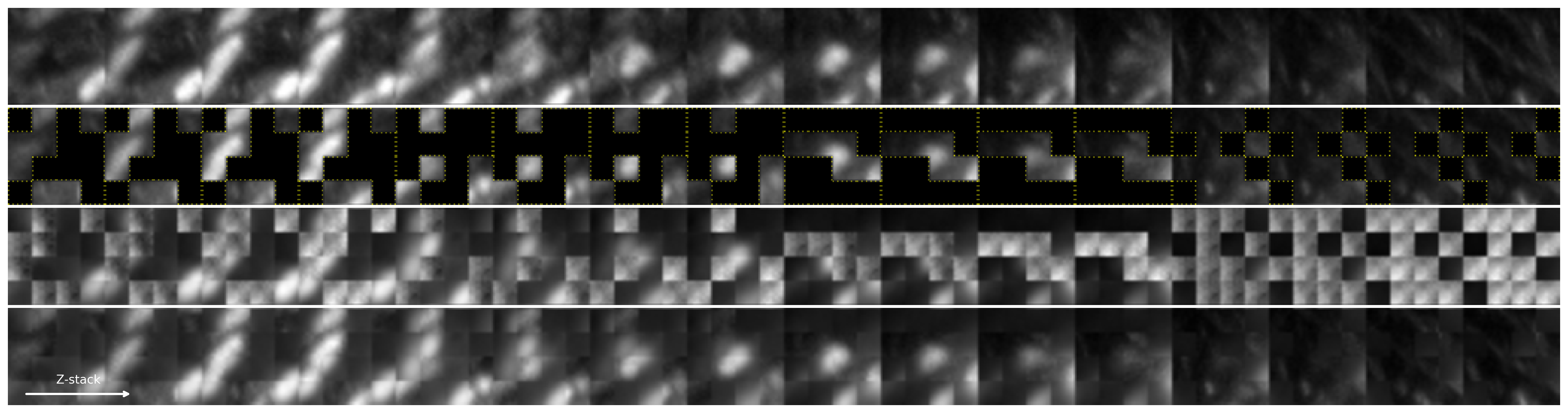}\\[0.5em]
\includegraphics[width=0.98\textwidth]{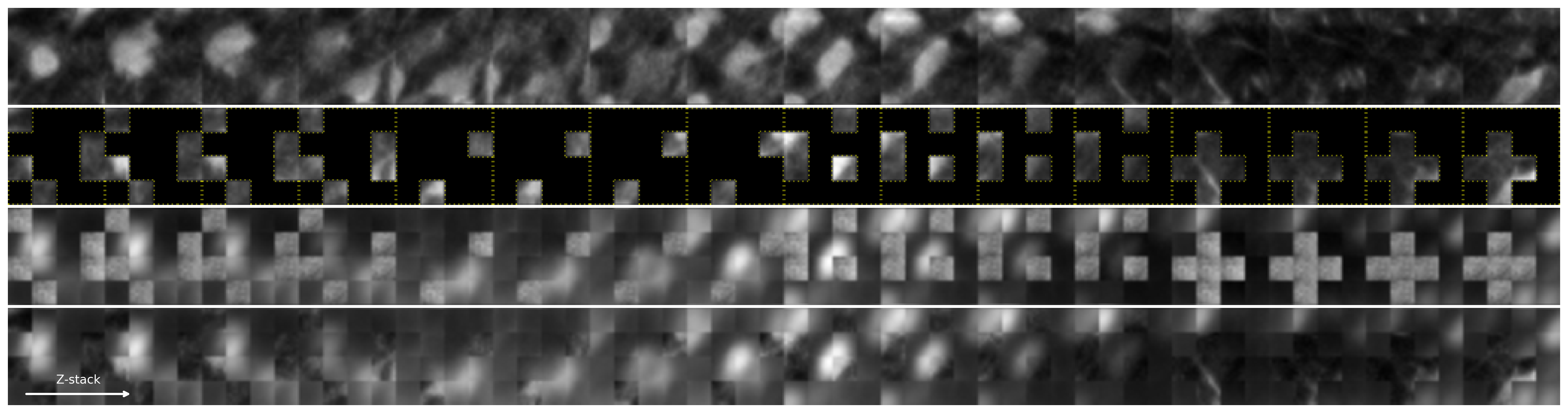}
\caption{\textbf{PV reconstruction with varying mask ratios.} Sparse interneurons with distinct morphology. 16×32×32 crop, 4×8×8 patch.}
\label{fig:mae_recon_pv}
\end{figure}

\begin{figure}[H]
\centering
\includegraphics[width=0.98\textwidth]{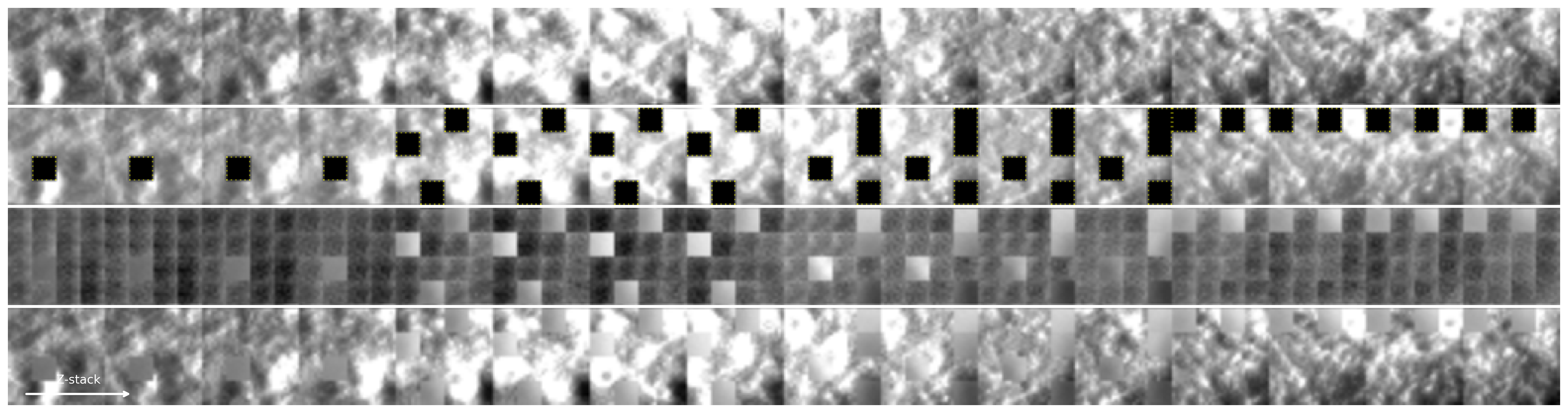}\\[0.5em]
\includegraphics[width=0.98\textwidth]{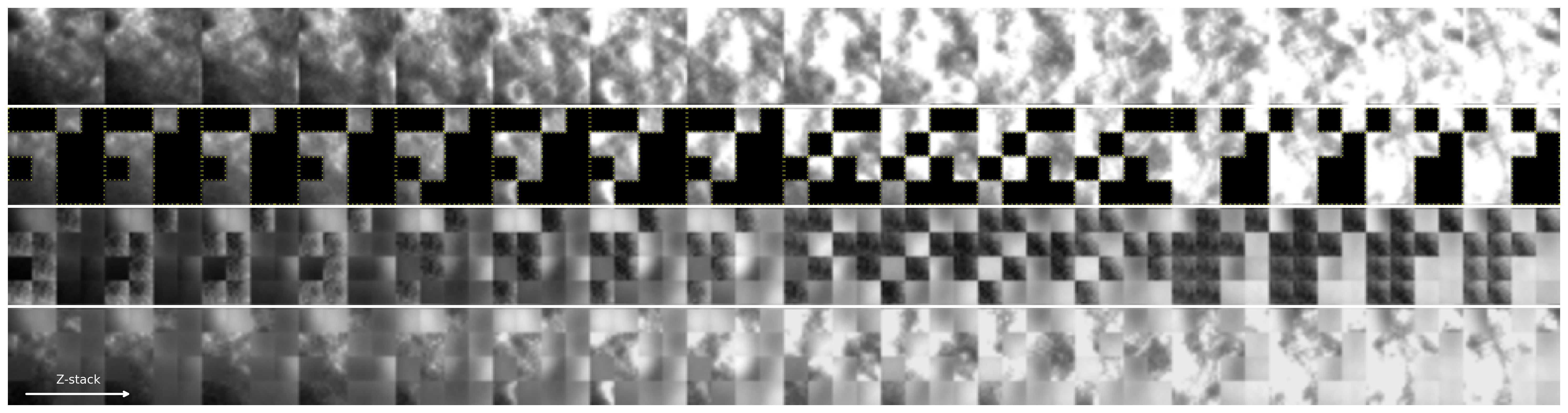}\\[0.5em]
\includegraphics[width=0.98\textwidth]{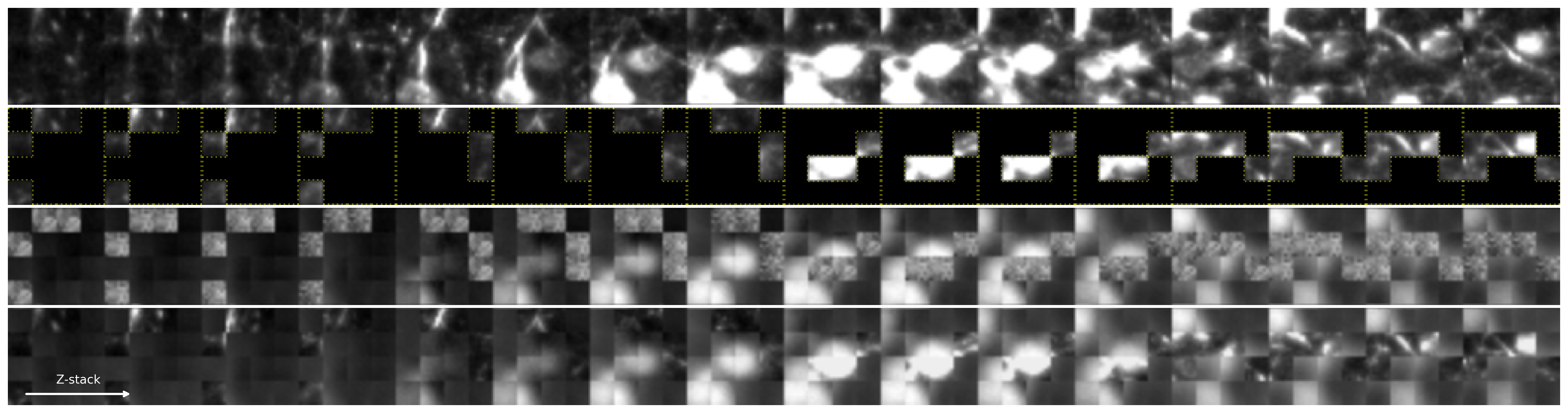}
\caption{\textbf{TH reconstruction with varying mask ratios.} Elongated dopaminergic neuron morphology. 16×32×32 crop, 4×8×8 patch.}
\label{fig:mae_recon_th}
\end{figure}

\begin{figure}[H]
\centering
\includegraphics[width=0.98\textwidth]{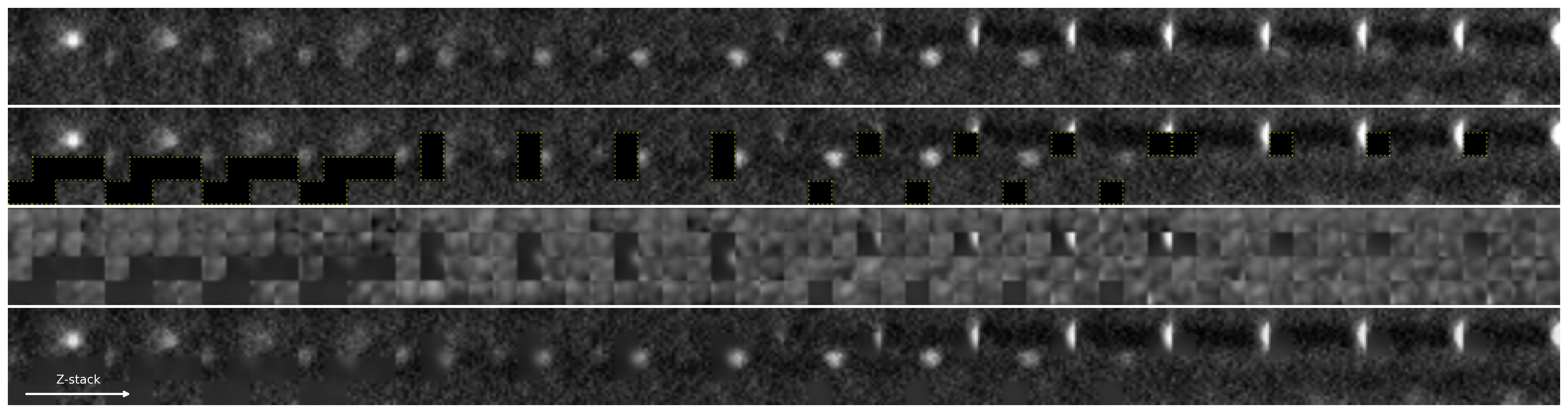}\\[0.5em]
\includegraphics[width=0.98\textwidth]{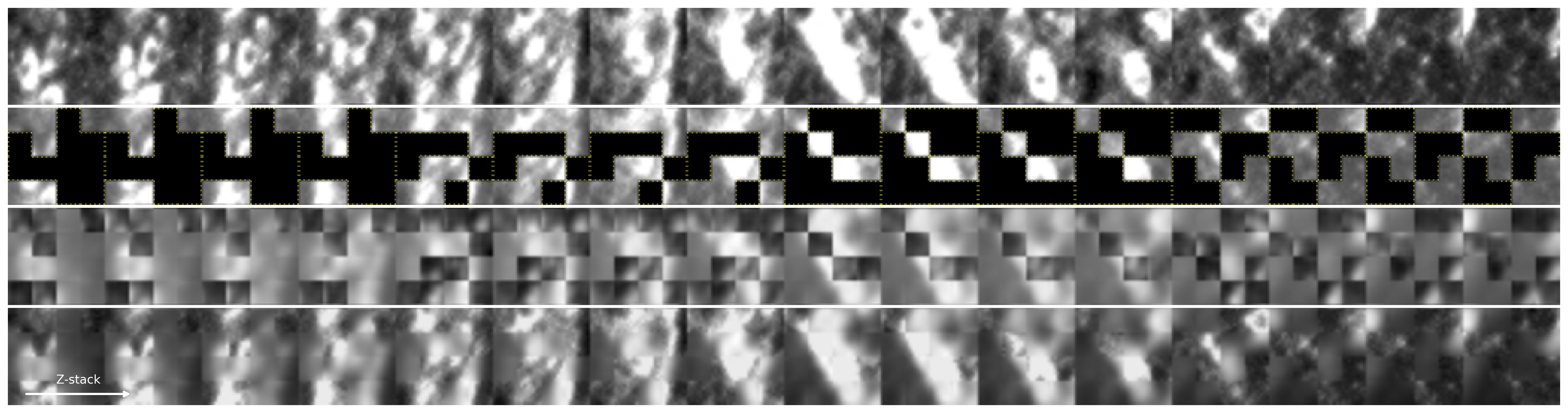}\\[0.5em]
\includegraphics[width=0.98\textwidth]{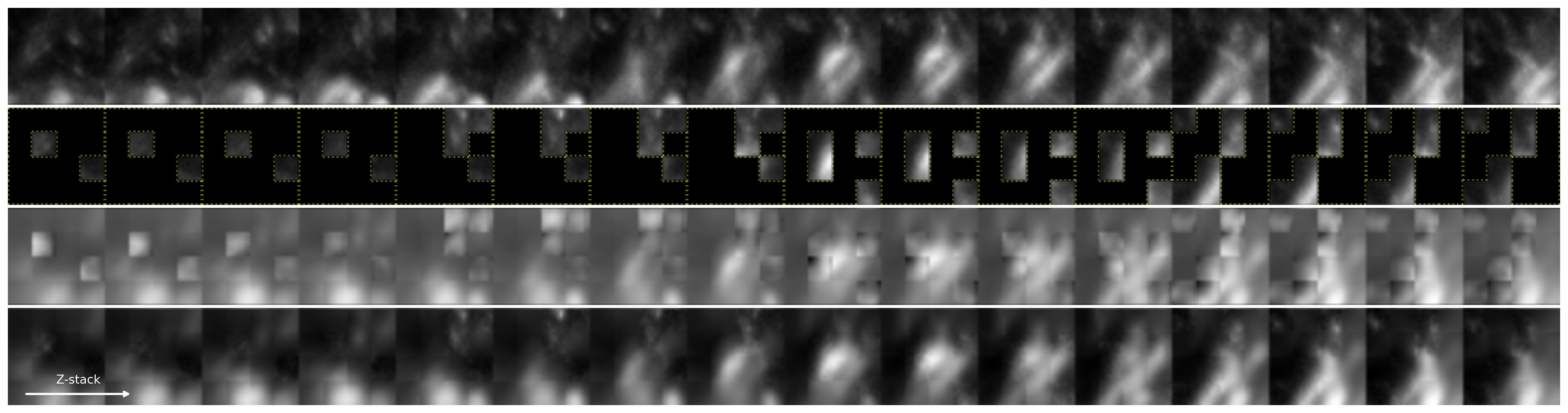}
\caption{\textbf{All-markers universal model reconstructions.} Training on diverse cell types (60k patches from all 6 markers) produces a generalizable encoder maintaining quality across morphologies. 16×32×32 crop, 4×8×8 patch.}
\label{fig:mae_recon_all}
\end{figure}

Representative reconstruction examples from the 3D-MAE training (\Cref{fig:mae_recon_neun_masks}, \Cref{fig:mae_recon_neun_crops}, \Cref{fig:mae_recon_iba1}, \Cref{fig:mae_recon_cfos}, \Cref{fig:mae_recon_gfap}, \Cref{fig:mae_recon_pv}, \Cref{fig:mae_recon_th}, \Cref{fig:mae_recon_all} ) illustrate the model’s ability to reconstruct cellular morphology from masked patches. Each figure shows all Z-slices concatenated horizontally, as indicated by the arrow, and arranged in four rows: the original volume, the masked input with yellow dotted borders marking masked areas, the reconstructed output, and the merged view combining visible and reconstructed regions.

Across markers, NeuN exhibits the sharpest reconstructions owing to its compact and uniform morphology, while IBA1 preserves fine ramified processes despite their complexity. GFAP shows some blurring along elongated processes but retains overall structural continuity. The all-markers model successfully reconstructs a wide range of morphologies, demonstrating good generalization. Lower mask ratios (e.g., 0.15) help preserve spatial context for dense cell types, whereas moderate ratios (e.g., 0.55) provide more effective regularization for sparsely distributed markers.

\subsection{CANVAS dataset structure}
Each cell type contains a full brain volume as sequential TIFF Z-slices and ground truth annotations for six ROIs split into train and test sets, while test sets are not publicly announced. The compressed .tar files are also available for each dataset.

\begin{minipage}{\columnwidth}
\DTsetlength{0.15em}{0.8em}{0.2em}{0.4pt}{1.6pt}
\footnotesize
\label{appendix_data_structure}
\dirtree{%
.1 CANVAS.
.2 [Marker]\_[BrainID] (e.g., IBA1\_brain\_11).
.3 [Marker]\_[BrainID].tar.
.3 image.
.4 344170\_392340\_000000\_ch2.tiff.
.4 344170\_392340\_000010\_ch2.tiff.
.4 \ldots\ (sequential Z-slices).
.3 label.
.4 TRAIN.
.5 [Marker]\_[brain]\_region1\_gt.csv.
.5 [Marker]\_[brain]\_region2\_gt.csv.
.5 [Marker]\_[brain]\_region3\_gt.csv.
.4 TEST (not publicly available).
.3 LICENSE.md.
}
\end{minipage}

\onecolumn

\end{document}